\documentclass[10pt,twocolumn,letterpaper]{article}

\usepackage{cvpr}              
\usepackage{amsmath}
\definecolor{cvprblue}{rgb}{0.21,0.49,0.74}
\usepackage[pagebackref,breaklinks,colorlinks,allcolors=cvprblue,urlcolor=magenta]{hyperref}

\def\paperID{2639} 
\def\confName{CVPR}
\def\confYear{2025}

\title{F\(^3\)OCUS - Federated Finetuning of Vision-Language Foundation Models with Optimal Client Layer Updating Strategy via Multi-objective Meta-Heuristics}
\author{
Pramit Saha$^1$ \and Felix Wagner$^1$ \and Divyanshu Mishra$^1$ \and Can Peng$^1$ \and Anshul Thakur$^1$ \and David A. Clifton$^{1,2}$ \and Konstantinos Kamnitsas$^{1,3,4}$ \and J. Alison Noble$^1$ \and \\
$^1$University of Oxford, $^2$Oxford-Suzhou Centre for Advanced Research, \\
$^3$Imperial College London, $^4$University of Birmingham \\
{\tt\small pramit.saha@eng.ox.ac.uk}
}

\begin{document}
\maketitle
\begin{abstract}
Effective training of large Vision-Language Models (VLMs) on resource-constrained client devices in Federated Learning (FL) requires the usage of parameter-efficient fine-tuning (PEFT) strategies. To this end, we demonstrate the impact of two factors \textit{viz.}, client-specific layer importance score that selects the most important VLM layers for fine-tuning and inter-client layer diversity score that encourages diverse layer selection across clients for optimal VLM layer selection. We first theoretically motivate and leverage the principal eigenvalue magnitude of layerwise Neural Tangent Kernels and show its effectiveness as client-specific layer importance score. Next, we propose a novel layer updating strategy dubbed \textbf{F$^3$OCUS} that jointly optimizes the layer importance and diversity factors by employing a data-free, multi-objective, meta-heuristic optimization on the server. We explore 5 different meta-heuristic algorithms and compare their effectiveness for selecting model layers and adapter layers towards PEFT-FL. Furthermore, we release a new MedVQA-FL dataset involving overall 707,962 VQA triplets and 9 modality-specific clients and utilize it to train and evaluate our method. Overall, we conduct more than 10,000 client-level experiments on 6 Vision-Language FL task settings involving 58 medical image datasets and 4 different VLM architectures of varying sizes to demonstrate the effectiveness of the proposed method. 
\end{abstract}
\section{Introduction}
\label{sec:introduction}

\begin{figure}
    \centering
\includegraphics[width=1.0\columnwidth]{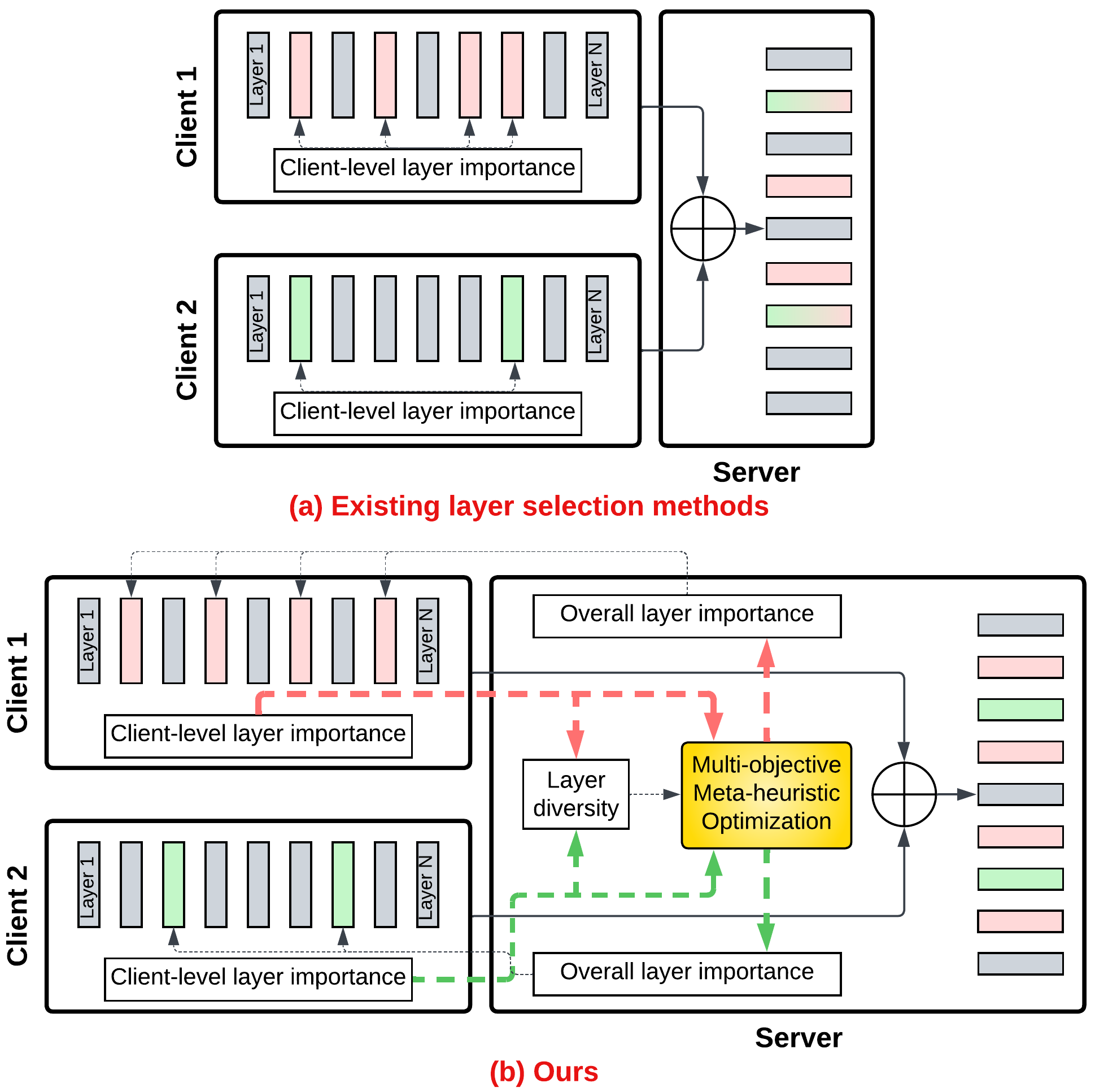}
    \caption{Distinction of our approach from prior works. (a) illustrates vanilla layer-selection process, which only selects parameter subsets based on the local client data without considering the requirements of the other clients. (b) depicts our approach, \textbf{F\textsuperscript{3}OCUS}, which refines the client-specific layer selection by jointly maximizing overall client-specific importance score and layer selection diversity score across clients.}
\label{fig:11}
\end{figure}

Large Vision-Language Models (VLMs) have made significant advancements in multi-modal learning, excelling in tasks like Visual Question Answering (VQA) \cite{bordes2024introductionvisionlanguagemodeling,li2021align,kim2021vilt,liu2024improved,li2023blip}. Their effectiveness stems from their extensive parameters often reaching millions or billions, allowing them to learn complex representations of image and text data. Fine-tuning these models with task-specific data is crucial for adapting them to specialized applications. However, gathering diverse training data centrally is challenging, especially in fields like healthcare, where strict privacy regulations prevent data aggregation across different centers.
To address the privacy concerns, Federated Learning (FL) \cite{mcmahan2017communication,li2020federated,karimireddy2020scaffold,acar2021federated}
allows models to be trained directly on local devices, such as in healthcare clinics, without sharing sensitive data. Yet, fine-tuning large models locally is difficult due to limited computational power and smaller datasets, which hinders VLM adaptation. Balancing privacy with these resource limitations requires innovative solutions like Parameter-Efficient Fine-Tuning (PEFT) that fine-tunes either selected model parameters or added parameters while keeping the original model fixed \cite{hu2022lora,rebuffi2018efficient,lian2022scaling,ben-zaken-etal-2022-bitfit,frankle2021trainingbatchnormbatchnormexpressive,touvron2022three,lester2021power,VPT_jia,li2021prefix}. Combined with FL, these offer a privacy-preserving and resource-efficient strategy for training large models collaboratively across multiple clients, particularly in computationally constrained settings.

\begin{figure}
    \centering
\includegraphics[width=1.0\columnwidth]{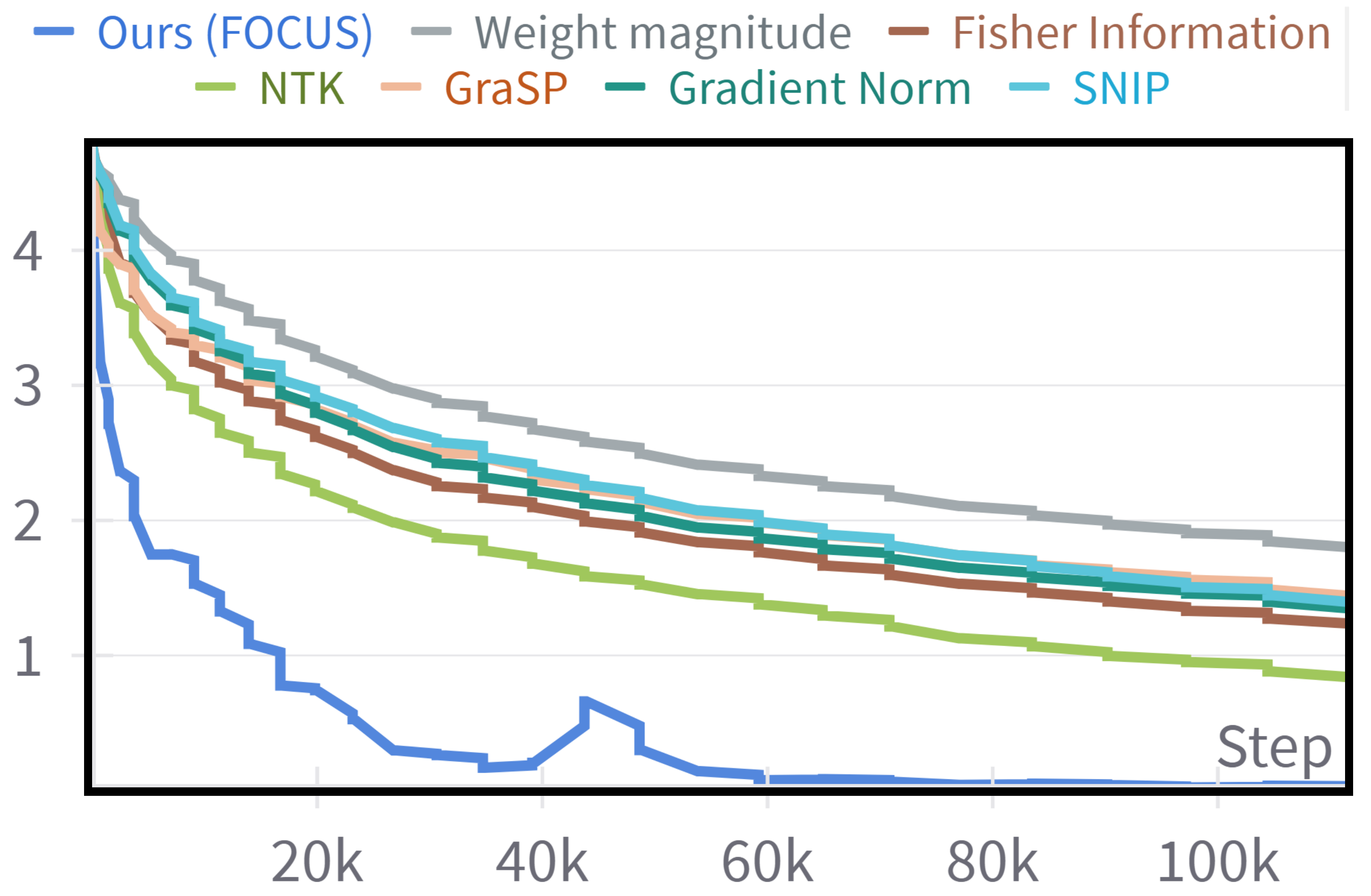}
    \caption{Loss convergence of layer selection methods. The gap between the client-specific NTK and FOCUS demonstrates the importance of our multi-objective meta-heuristic optimization.}
\label{fig:11}
\end{figure}

Previous research \cite{chen2022fedtune, sun2022exploring,yang2024exploring,yu2023federated,zhuang2023foundation,zhang2024towards} has mostly focused on naive combination of centralized finetuning methods with FedAvg \cite{mcmahan2017communication}. However, these works are primarily confined to single-modalities, addressing either visual or textual inputs independently. Besides, they do not consider the diverse characteristics and capacities of individual clients and typically assume homogeneous computational resources across all clients \cite{saha2024fedpia,chen2024feddat} which is not applicable in most real-world collaborative settings. Hence, some clients either under-utilize the available resources or are unable to participate due to lack of compute.
Our flexible layer selection PEFT-FL framework effectively addresses these issues.

The naive combinations of centralized selective PEFT methods \cite{ruckle2020adapterdrop,sung2021training,lee2018snip,lubanagradient,wang2020picking,lee2022surgical} and FL only consider local client data and task for selecting parameter subsets without considering other client requirements. This is especially problematic for FL clients facing challenges like heterogeneous modalities and computes, domain shifts, and statistical heterogeneity. In such cases, a poorly chosen client-specific configuration can not only slow down the overall convergence (see Fig. 2), but may also perform worse than training each client independently. Achieving the best performance requires a tailored approach that jointly considers client-specific as well as global optimization requirements.

To this end, we present a novel framework called \textbf{F\textsuperscript{3}OCUS} (\textbf{F}ederated \textbf{F}inetuning of \textbf{F}oundation Models with \textbf{O}ptimal
\textbf{C}lient-specific Layer \textbf{U}pdating \textbf{S}trategy) to improve layer selection by considering both local and global FL characteristics while respecting the client-specific resource constraints. 
We propose a two-step ``define and refine" procedure \textbf{at the beginning of every round}: (a) \textbf{client-level strategy}, that defines layer importance scores based on the principal eigenvalue of layerwise Neural Tangent Kernel (LNTK) and (b) \textbf{server-level strategy}, that refines client-specific layer selection by maximizing the overall client-specific importance scores while simultaneously minimizing the variance of the histogram of layer selections across clients, thereby promoting a more uniform distribution of layer participation. Our method (see Figs. 1 \& 4) provides the clients with a flexible and dynamic solution for selecting layers where each client can specify their computational budgets, while ensuring faster convergence (see Fig. 2). In order to showcase the effectiveness of \textbf{F\textsuperscript{3}OCUS}, we conduct over \textbf{10,000} client-level experiments under \textbf{6} Vision-language FL task settings using \textbf{4} VLMs and 58 medical image datasets that involve \textbf{4} types of heterogeneities based on data, modality, device, and task. Our primary contributions can be summed up as follows:

\begin{itemize}

\item \textbf{Dataset contribution:} We release Ultra-MedVQA, \textbf{the largest medical VQA dataset to date,} consisting of \textbf{707,962} VQA triplets including \textbf{9} different modalities and \textbf{12} distinct anatomies, covering diverse open-ended and closed questions related to modality, tissue type, image view, anatomy, and disease (see Tab. 1 for comparison). 

\item \textbf{Technical contributions:}
We theoretically motivate and propose \textbf{F\textsuperscript{3}OCUS}, a new selective layer fine-tuning strategy. We introduce LNTK-based client-level layer selection and server-level multi-objective meta-heuristic optimization that jointly optimizes client-specific layer importance score and inter-client layer diversity score. We theoretically motivate and analyze the effectiveness of our layer selection strategy and prove its convergence. 

\item \textbf{Empirical contributions:} Unlike previous works, we consider more constrained and realistic client settings involving data, modality, task, and device heterogeneity. We conduct comprehensive evaluations of \textbf{F\textsuperscript{3}OCUS} with 4 VLMs of varying sizes in diverse FL settings for tuning selective layers. 
Consequently, we present detailed insights into \textbf{F\textsuperscript{3}OCUS}'s performance improvements through: (a) analysis of client and server-based layer rank and importance score computation during training and (b) evaluation of different meta-heuristic optimization algorithms on the server \textit{viz}., Genetic Algorithm, Artificial Bee Colony, Ant Colony Optimization, Simulated Annealing, and Swarm Particle Optimization.

\end{itemize}

\begin{table}[t]
\centering
\caption{\# modalities (M) and VQA triplets in different datasets}
\scalebox{.6}{
\begin{tabular}{l|cccccc}
\hline
 & \textbf{VQA-RAD \cite{lau2018dataset}} & \textbf{SLAKE \cite{liu2021slake}}& \textbf{Path-VQA \cite{he2020pathvqa}} & \textbf{VQA-Med \cite{BenAbacha2020VQAMed}}  & \textbf{OmniMedVQA \cite{Hu_2024_CVPR}} & \textbf{Ours} \\
\hline
\textbf{\# M} & 3 & 3 & 2 & 5 & 12 & 9 \\
\textbf{\# VQA} & 3515 & 14028 & 32799 & 5500 & 127995 & 707962\\
\hline
\end{tabular}}
\label{table:comparison}
\end{table}

\begin{figure*}
    \centering
\includegraphics[width=2\columnwidth]{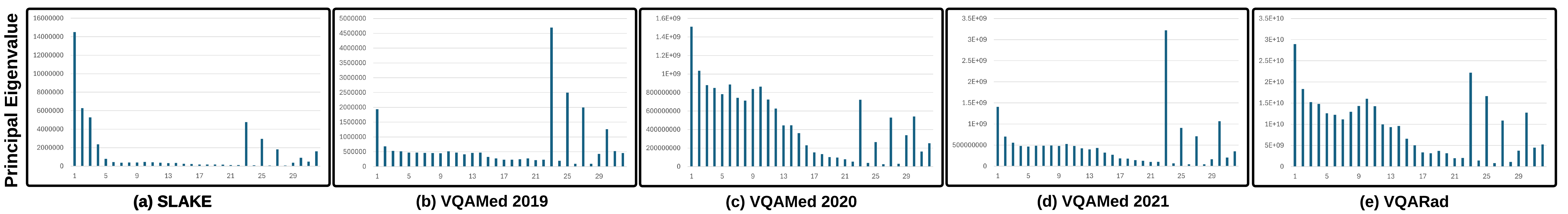}
    \caption{Visualization of principal eigenvalue magnitudes of LNTK (see $\S 4.1$) for computing layer importance score of LLaVA-1.5-7b} 
\label{fig:11}
\end{figure*}
\section{Background and Related works}

\textbf{Federated Learning (FL)}
FL enables various clients to collaboratively train models in a decentralized manner without sharing local data. The classical FL framework, FedAvg \cite{mcmahan2017communication}, offers a practical method for model aggregation. 
Several modifications have emerged to address the adverse impact of data heterogeneity in FL \cite{li2020federated,karimireddy2020scaffold,acar2021federated,saha2023rethinking,saha2024examiningmodalityincongruitymultimodal,wagner2023post,wagner2024feasibility,hernandez2024review}.

\noindent
\textbf{Centralized selective fine-tuning:} Various methods have been explored for selecting subsets of parameters for fine-tuning foundation models in centralized training. These include optimizing non-structured mask matrices \cite{leemixout,zhang2022fine,xu2021raise,lee2019would,kovaleva2019revealing,zaken2021bitfit,zhang2023crash,shen2021partial}, employing layer-wise selection \cite{kovaleva2019revealing,lee2019would,lee2022surgical,kaplun2023less} and pruning methods \cite{rachwan2022winning,wang2020picking,lee2018snip,zhao2024pruning,li2020efficient}.

\noindent
\textbf{Federated selective fine-tuning:} Recent research has adapted these selective PEFT methods for FL \cite{nguyen2022begin,chen2022fedtune,hilmkil2021scaling,zhang2022fine}. Specifically, studies by \cite{lee2023layer,dun2022resist} explore layer-wise network decomposition to facilitate selective model fine-tuning on client devices. Partial model personalization algorithms \cite{chen2023efficient,pillutla2022federated} aim to train tailored subnetworks on clients to improve local models. However, these studies do not provide adaptive or dynamic layer selection strategies that consider the diverse characteristics of clients. Unlike prior works, we account for client-specific differences in resources and data distributions while also considering the global optimization requirements to perform selective layer fine-tuning.

\begin{figure*}
    \centering
\includegraphics[width=2.0\columnwidth]{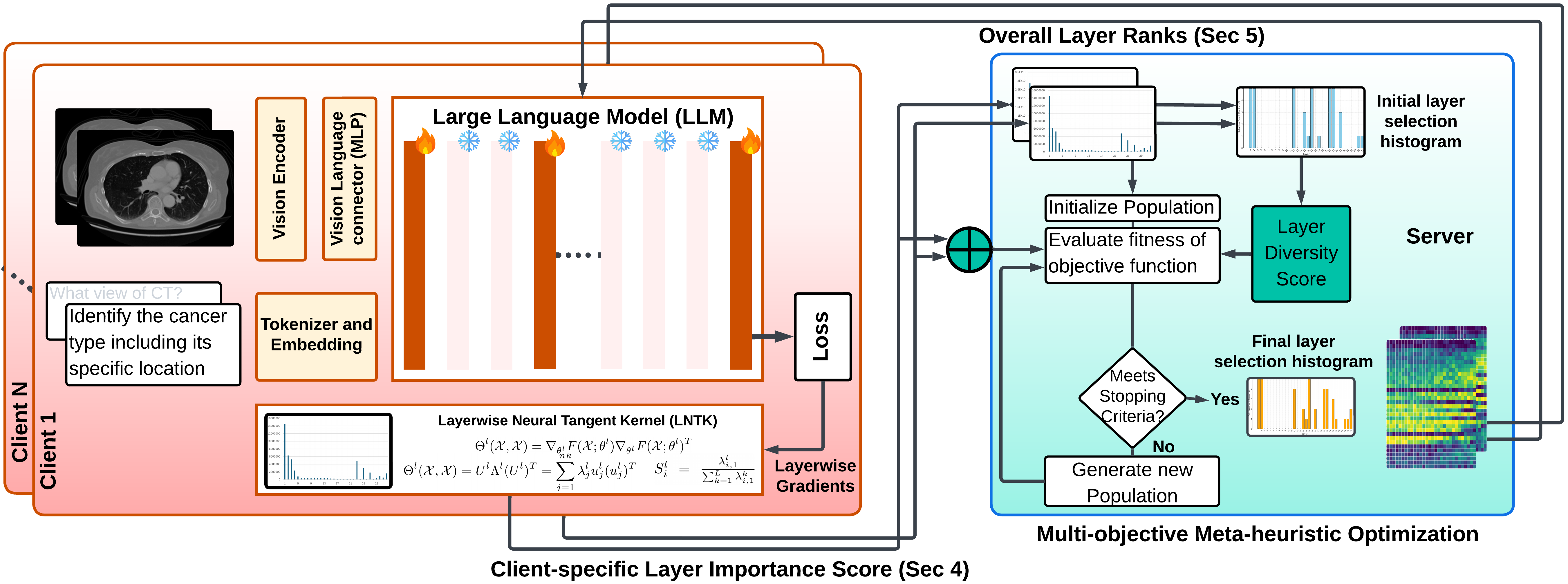}
    \caption{Overview of our layer selection strategy, \textbf{F\textsuperscript{3}OCUS}. Each client sends layer importance scores based on the principal eigenvalue of LNTK to the server. The server refines client-specific layer selection by maximizing the cumulative client-specific importance scores while simultaneously minimizing the variance of the histogram of layer selections across clients. It sends the revised layer ranks back.} 
\label{fig:11}
\end{figure*}

\section{Problem Formulation}

Consider an FL system with a central server and $N$ clients, represented by $\mathcal{N} = \{1, \ldots, N\}$. Each client has its own private dataset $\mathcal{D}_i$, containing $d_i = |\mathcal{D}_i|$ data points. The server contains a pre-trained foundation model parameterized by $\theta \in \mathbb{R}^P$, comprising $L$ layers, indexed by $\mathcal{L} = \{1, 2, \ldots, L\}$. The server's objective is to fine-tune this model based on the clients' data $\{\mathcal{D}_i\}_{i \in \mathcal{N}}$ without directly accessing these datasets. The learning goal is formalized as:
\begin{equation}
\small
\min_{\theta \in \mathbb{R}^P} F(\theta) = \sum_{i=1}^{N} \alpha_i F_i(\theta),
\end{equation}
where $\alpha_i = \frac{d_i}{\sum_{j=1}^{N} d_j}$ denotes relative sample size, and $F_i(\theta) = \frac{1}{d_i} \sum_{B_i \in \mathcal{D}_i} F_i(\theta; B_i)$ represents local training objective for client $i$, with $F_i(\theta; B_i)$ being the (potentially non-convex) loss function of model $\theta$ on data batch $B_i$. The FL training process proceeds over $T$ rounds. In each round $t \in [T]$, the server selects a subset of clients $\mathcal{S}_t$ and distributes the updated global model $\theta_t$ to them for training.

Due to resource constraints, instead of the entire model, clients update only a subset of layers during local training. Each client-specific masking vector can be denoted as $m_{i,t} \in \{0,1\}^L$, where $m^l_{i,t} = 1$ if layer $l$ is selected for training in round $t$, and $m^l_{i,t} = 0$ otherwise. Thus, the set of selected layers for client $i$ at round $t$ is denoted as $\mathcal{L}_{i,t} = \{l \in \mathcal{L} \mid m^l_{i,t} = 1\}$, and the union of all selected layers across clients in round $t$ is $\mathcal{L}_t = \bigcup_{i \in \mathcal{S}_t} \mathcal{L}_{i,t}$. 

Clients initialize their local models using global model from the server, $\theta_{i,t}$, and perform $\tau$ steps of local training using mini-batch SGD. For each local step $k \in [\tau]$, client $i$ samples a batch of data $B_{i,t}$ and calculates gradients for the selected layers as:
\begin{equation}
\small
\sum_{l \in \mathcal{L}^t_i} G_{i,l}(\theta^{k}_{i,t}; B^{k}_{i,t}) = \sum_{l \in \mathcal{L}^t_i} \nabla_l F_i(\theta_{i,t}; B_{i,t}),
\end{equation}

where $\nabla_l F(\theta)$ denotes the gradient of $F(\theta)$ with respect to the parameters of layer $l$. After computing the gradients, the local model is updated with learning rate $\eta$ as:
\begin{equation}
\small
\theta^{k}_{i,t} = \theta^{k-1}_{i,t} - \eta \sum_{l \in \mathcal{L}^t_i} G_{i}^{l}(\theta^{k-1}_{i,t}; B^{k-1}_{i,t}), \quad \forall k \in \{1, 2, \ldots, \tau\},
\end{equation}
The accumulated weight update in one local round is:
\begin{equation}
\small
    \delta_{i,t} = \frac{1}{\eta} (\theta_{i,t}^0 - \theta_{i,t}^\tau) =  \sum_{k=0}^{\tau-1}\sum_{l \in \mathcal{L}^t_i} G_{i}^{l}(\theta^{k}_{i,t}; B^{k}_{i,t})
\end{equation}
Accumulated update after federated aggregation on server:
\begin{equation}
\small
    \delta_{t} = \sum_{i \in S_t}\sum_{k=0}^{\tau-1}\sum_{l \in \mathcal{L}^t_i} \alpha_i G_{i}^{l}(\theta^{k}_{i,t}; B^{k}_{i,t})
\end{equation}
\section{Client-level layer importance}
\subsection{Layerwise Neural Tangent Kernel (LNTK)}
We first motivate the usage of LNTK towards prioritizing selected layers for client-specific fine-tuning. To capture the model training dynamics, consider the evolution of $\theta_t$ and neural network function $f$ over input instances $\mathcal{X} \subset \mathcal{D}_i$:
\begin{equation}
\small
\begin{aligned}
\dot{\theta}_t &= -\eta \nabla_{\theta} f(\mathcal{X}; \theta_t)^T \nabla_{f} F (\mathcal{X}; \theta_t), \\
\dot{f} &= \nabla_{\theta} f(\mathcal{X}; \theta_t) \dot{\theta}_t 
= -\eta \nabla_{\theta} f(\mathcal{X}; \theta_t)\nabla_{\theta} f(\mathcal{X}; \theta_t)^T \nabla_{f} F (\mathcal{X}; \theta_t), \\
\dot{f} &= -\eta \Theta(\mathcal{X}, \mathcal{X}) \nabla_{f} F (\mathcal{X}; \theta_t)
\end{aligned}
\end{equation}

\noindent
where $F$ is training loss. NTK matrix $\Theta_t(\mathcal{X}, \mathcal{X})$ at time $t$:
\begin{equation}
\small
\Theta_t(\mathcal{X}, \mathcal{X}) \triangleq \nabla_{\theta} f(\mathcal{X}; \theta_t) \nabla_{\theta} f(\mathcal{X}; \theta_t)^T \in \mathbb{R}^{n \times n}. 
\end{equation}

\noindent
The integral NTK of a model \cite{jacot2018neural} can be computed as the sum of layerwise NTK (LNTK), where LNTK \cite{shi2024train} is:
\begin{equation}
\small
\Theta^l(\mathcal{X}, \mathcal{X}) = \nabla_{\theta^l} f (\mathcal{X}; \theta^l) \nabla_{\theta^l} f(\mathcal{X}; \theta^l)^T,
\end{equation}

\noindent
where $\nabla_{\theta^l} f(\mathcal{X}; \theta^l)$ denotes the Jacobian of $f$ at input points $\mathcal{X}$ with respect to the $l$-th layer parameters $\theta^l$. The integral NTK can be expressed as: 

\begin{figure*}[t]
    \centering
\includegraphics[width=2.0\columnwidth]{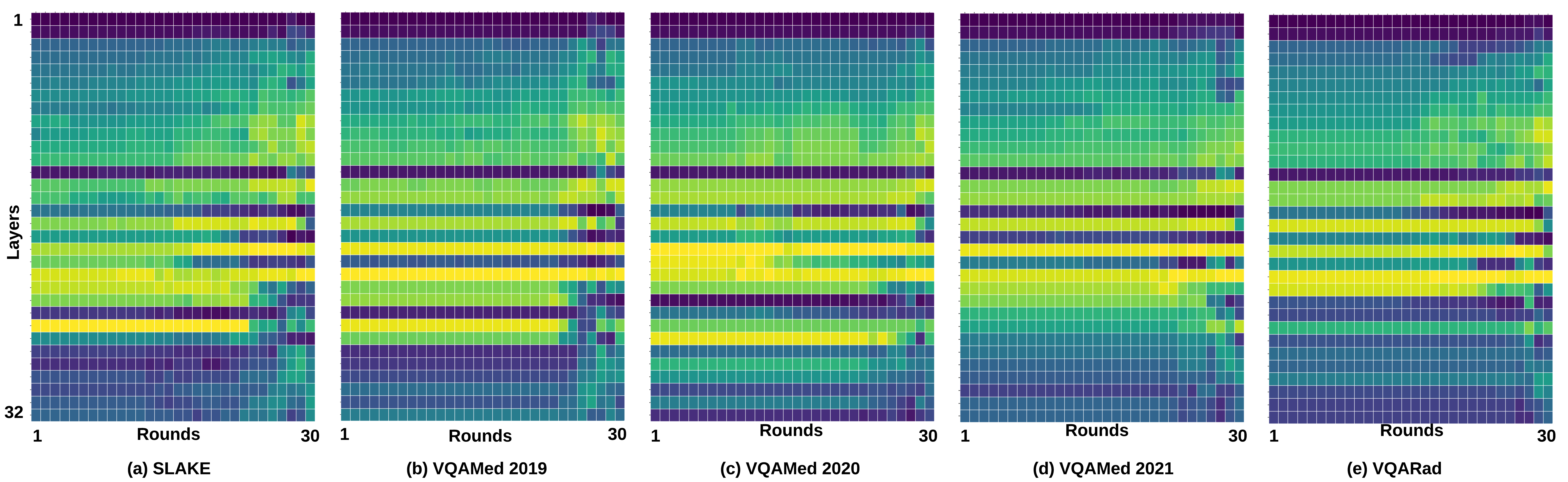}
    \caption{Visualization of layer ranks of LLaVA-1.5 across rounds in different clients based on LNTK. Darker color implies higher rank.}    
\label{fig:11}
\end{figure*}
\begin{equation}
\small
\begin{aligned}
\Theta(\mathcal{X}, \mathcal{X}) 
&\overset{(i)}{=} \sum_{l=1}^L \nabla_{\theta^l} f(\mathcal{X}) \nabla_{\theta^l} f(\mathcal{X})^T \\
&\overset{(ii)}{=} \sum_{l=1}^L \sum_{\theta_p \in \theta^l} \nabla_{\theta_p} f(\mathcal{X}) \nabla_{\theta_p} f(\mathcal{X})^T
&\overset{(iii)}{=} \sum_{l=1}^L \Theta^l(\mathcal{X}, \mathcal{X})
\end{aligned}
\end{equation}
where (i) decomposes the matrix multiplication into the sum of vector multiplications; (ii) gathers addends by each module; and (iii) follows the definition of the LNTK. Since $\Theta^l(\mathcal{X}, \mathcal{X})$ is a positive semi-definite real symmetric matrix, we perform an eigen-decomposition of LNTK as:
\begin{equation}
\small
\Theta^l(\mathcal{X}, \mathcal{X}) = U^l \Lambda^l (U^l)^T = \sum_{j=1}^{nk} \lambda^l_j u^l_j (u^l_j)^T
\end{equation}
where $\Lambda^l = \operatorname{diag}(\lambda^l_1, \lambda^l_2, \dots, \lambda^l_{nk})$ contains eigenvalues $\lambda^l_j$ of $\Theta^l(\mathcal{X}, \mathcal{X})$, and each $\lambda^l_j \geq 0$. The mean output of the $l$-th layer in the eigenbasis of the LNTK can be described by:
\begin{equation}
\small
(U^lE[\mathbf{f}^l(\mathcal{X})])_j = \left(I - e^{-\eta \lambda_j^l t}\right)(U^ly^l)_j
\end{equation}
where $\mathbf{f}^l(\mathcal{X})$ and $y^l$ are actual and target $l$-th layer outputs. This formulation demonstrates that the convergence behavior of a layer is largely influenced by the eigenvalues $\lambda_j^l$ of LNTK. Let $\lambda^l_{1} \geq \lambda^l_{2} \geq \cdots \geq \lambda^l_{n}$ denote the eigenvalues of $\Theta^l(\mathcal{X}, \mathcal{X})$, where $\lambda^l_{1}$ is the principal eigenvalue.
In particular, $\lambda_1^l$ plays a dominant role in the convergence dynamics \cite{bowman2023brief}. When using the largest possible layer-specific learning rate, $\eta^l \sim \frac{2}{\lambda_1^l}$, the training process aligns most strongly with the direction associated with the principal eigenvalue as the NTK eigenvectors corresponding to the principal eigenvalue are learned quicker due to spectral bias \cite{rahaman2019spectral,bowman2023brief,cao2019towards,shi2024train}.
This motivates the use of the principal eigenvalue $\lambda^l_{1}$ in our client-specific layer importance score, as it represents the maximum alignment of each layer's parameter space with the client's data distribution. This provides a principled basis for prioritizing these layers.

For a step of gradient descent, the loss reduction can be characterized by the directional derivative of the loss:
\begin{equation}
\small
\begin{aligned}
\Delta F 
&\overset{(i)}{=} \lim_{\epsilon \to 0} \frac{F(\theta + \epsilon \nabla_{\theta} F(\theta)) - F(\theta)}{\epsilon} 
\overset{(ii)}{\approx} \nabla_{\theta} F(\theta)^T \nabla_{\theta} F(\theta) \\
&\overset{(iii)}{=} \nabla_{Z} F(\theta)^T (\nabla_{\theta} F(\theta)^T \nabla_{\theta} F(\theta)) \nabla_{Z} F(\theta) \\
&\overset{(iv)}{=} \nabla_{Z} F(\theta)^T \left( \sum_{l=1}^L \Theta^l \right) \nabla_{Z} F(\theta) 
\overset{(v)}{=} \sum_{l=1}^L \sum_{j=1}^{nk} \lambda^l_j \left( (u^l_j)^T Y \right)^2 \\&\overset{(vi)} \approx \sum_{l=1}^L \lambda^l_1 \left( (u^l_1)^T Y \right)^2
\end{aligned}
\end{equation}

where (i) follows the definition of the directional derivative; (ii) follows the first-order Taylor expansion; (iii) applies the chain rule of derivatives; (iv) follows from Eq. (8); and (v) follows the eigen-decomposition of the layerwise NTK under the assumption of squared error loss. Assuming that the true labels align well with the top eigenvectors as discussed earlier, \textit{ i.e.}, $\left( (u^l_j)^T Y \right)^2$ is large for large $\lambda^l_j$, directional derivative of the loss function can be regarded as closely related to the eigenspectrum of the layerwise NTKs. Specifically, (vi) suggests that layers with higher principal eigenvalues contribute more significantly to loss reduction during training. Overall, Eq. 12 suggests that the loss decreases more rapidly along the eigenspaces corresponding to larger LNTK eigenvalues. Given that the principal eigenvalue $\lambda^l_1$ vary across layers as seen in Fig. 3, we propose to selectively fine-tune layers with larger $\lambda^l_1$ to achieve efficient learning with limited computational resources.

Therefore, we define the \textbf{client-specific layer importance score} $S^l_i = \frac{\lambda^l_{i, 1}}{\sum_{k=1}^L \lambda^k_{i, 1}}$ where the sum in the denominator normalizes the principal eigenvalue across all layers, ensuring that $S^l_i$ captures the relative importance of each layer for client $i$ in terms of its contribution to the model’s predictive capacity for that client's data distribution. This formulation prioritizes layers whose NTK principal eigenvalues dominate, indicating strong client-specific parameter alignment (see Figs. 3 \& 5). \textbf{See Algorithm in Suppl. \S A.}
\subsection{Convergence Analysis of LNTK}
We begin with some necessary assumptions following previous works \cite{wang2020tackling,karimireddy2020scaffold} and then introduce a set of assumptions to analyze the impact of the layers selected using LNTK:

\noindent
\textbf{Assumption 1: ($\gamma$-Smoothness)} There exists a constant $\gamma > 0$ such that for any $\theta, \phi \in \mathbb{R}^p$:
\begin{equation}
\small
\|\nabla F_i(\theta) - \nabla F_i(\phi)\|_2 \leq \gamma \|\theta - \phi\|_2, \quad \forall i \in \mathcal{N}.
\end{equation}

\noindent
\textbf{Assumption 2: (Unbiased and variance-bounded stochastic gradient)} The layerwise stochastic gradient \( G_{i,l}(\theta_t; B_{i,t}) \) computed on a randomly sampled data batch \( B_{i,t} \) serves as an unbiased estimate of the layerwise full-batch gradient:

\noindent
\begin{equation}
\small
\mathbb{E}_{B_{i,t}}[G_{i,l}(\theta_t; B_{i,t})] = \nabla F_{i,l}(\theta_t).
\end{equation}
\noindent

Besides, there exist constants $\sigma_l > 0, \forall l \in \mathcal{L}$ such that $\mathbb{E}_{B_{i,t}} \|G_{i}^{l}(\theta_t; B_{i,t}) - \nabla F_{i}^{l}(\theta_t)\|^2 \leq \sigma_l^2, \forall i \in \mathcal{N}$ and $\sum_{l \in \mathcal{L}} \sigma_l^2 \leq \sigma^2$.

\noindent
\textbf{Assumption 3: (Gradient Diversity)} The non-IID client data distribution causes diverse gradients. There exist constants $\kappa_l > 0, \forall l \in \mathcal{L}$ such that:\vspace{-2mm}
\begin{equation}
\small
\mathbb{E}_{B_{i,t}} [\|\nabla F(\theta_t) - \nabla F_{i}^{l}(\theta_t; B_{i,t})\|^2] \leq \kappa_l^2, \forall i \in \mathcal{N}.
\end{equation}


In contrast to the theoretical analysis for standard FL settings, our LNTK-based fine-tuning introduces three additional challenges: \textit{(i)} Each client only updates a subset of layers. Hence, the aggregated gradient is no longer an unbiased estimate of the local gradient $\nabla F_{i}(\theta_t)$, \textit{i.e.}, $\sum_{l \in \mathcal{L}_t} \nabla F_{i}^{l}(\theta_t) \neq \nabla F_{i}(\theta_t)$ 
where equality holds only if all layers are selected. 
\textit{(ii}) LNTK-based layer selection may vary across clients as seen in Fig. 5. The aggregated gradient of the selected layers is not equal to the gradient computed based on the global loss function \textit{i.e.,} $\sum_{i \in \mathcal{S}^t} \sum_{l \in \mathcal{L}_t}  \alpha_{i,t} \nabla_l F_i(\theta_t) \neq \sum_{l \in \mathcal{L}_t} \nabla_l F(\theta_t),$
where equality holds only if all clients select the same layers.

\noindent
\textit{(iii)} The gradient magnitudes in \textit{(i-ii)} vary across epochs.

In order to relate the aggregated and target gradient, we define a proxy layerwise loss $\psi_{i,l}$ optimized by clients  as:
\begin{equation}
\small
\psi_{i}^{l}(\theta_t) \triangleq \sum_{i \in \mathcal{S}_t} m_{i,t}^l\alpha_{i,t} F_{i}^{l}(\theta_t), \quad \text{s.t } \nabla_l \psi_{i}^{l}(\theta_t) = \delta_t
\end{equation}
Given assumptions 1-3, we formulate the convergence as:
\noindent
\textbf{Theorem 1} (Convergence of LNTK-based layer selection):
\begin{equation}
\small
\begin{aligned}
\min_{t \in [T]} \mathbb{E} \left[ \left\| \nabla F(\theta_t) \right\|_2^2 \right] \leq 
\frac{2}{(\eta - \gamma \eta^2) T} \Bigg[ \left[ F(\theta^0) - F(\theta^*) \right] \\ 
+ \gamma (\eta \sigma)^2 T 
+ \sum_{t=1}^{T} \left(\eta + \frac{1}{2 \gamma} - \gamma \eta^2 \right) \Bigg( \mathbb{E} \left[ \left\| \sum_{\ell \notin \mathcal{L}_t} \nabla_\ell F(\theta_t) \right\|^2 \right] \\
+ \sum_{\ell \in \mathcal{L}_t} \sum_{i \in \mathcal{N}} \alpha_{i,t} (1 - m_{i,t}^l)^2 k_l^2\Bigg) \Bigg] \quad \text{s.t } \theta^* = \arg \min_{\theta \in \mathbb{R}^p} F(\theta) 
\end{aligned}
\end{equation}


\noindent
\textbf{Remark 1:} With commonly chosen \( \eta = O\left(\frac{1}{\sqrt{T}}\right) \), RHS of (17) the last term $\to 0$ as \( T \to \infty \). So, LNTK-based PEFT-FL converges to a small neighbourhood of a stationary point of standard FL, maintaining a non-zero error floor. \textbf{See Suppl. \S B for complete proof and discussions.}

\section{Server-level Overall Layer Importance}
\begin{figure*}
    \centering
\includegraphics[width=2.0\columnwidth]{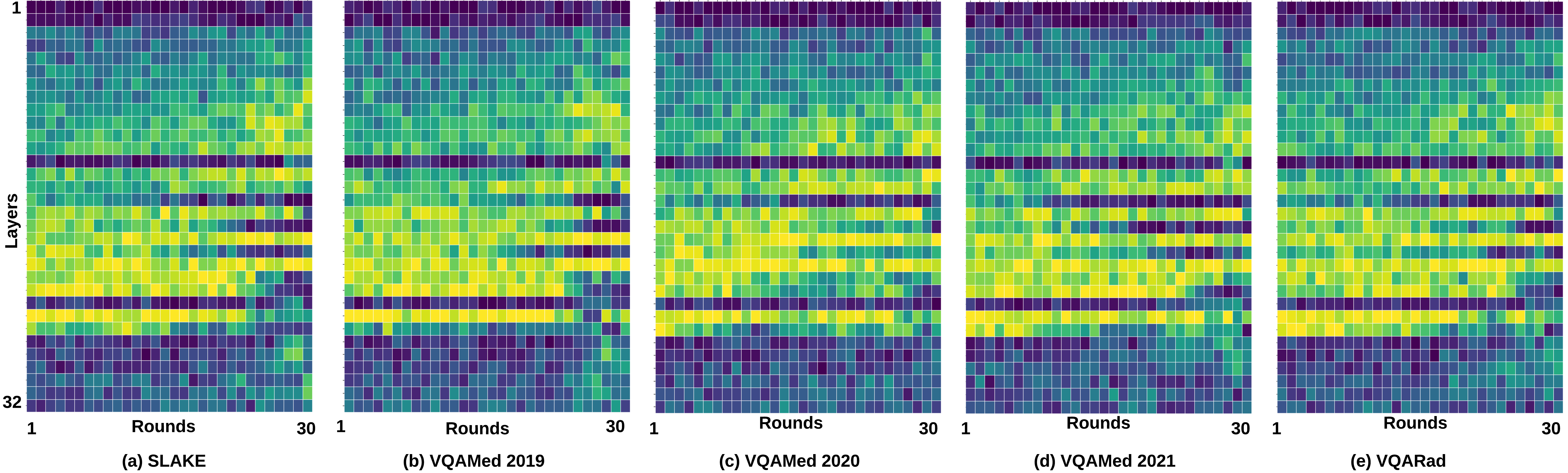}
    \caption{Visualization of refined layer ranks of LLaVA-1.5 based on \textbf{F\textsuperscript{3}OCUS}. Comparing it with Fig. 5 shows that the server-level meta-heuristic optimization refines the client-level layer ranks by increasing inter-client layer diversity at every round.} 
\label{fig:11}
\end{figure*}

\subsection{Theoretical Motivation}
To explicitly examine the impact of potential noise introduced by LNTK-based layer selection as well the variation of selection count across clients, we reformulate the convergence in Theorem 2 leveraging two additional assumptions:

\noindent
\textbf{Assumption 4: (Bounded stochastic gradient)}
The expected squared norm of stochastic gradients is bounded uniformly, \textit{i.e}., for constant $\sigma_g > 0$ and any $i,t$:
\begin{equation}
\small
\mathbb{E}_{B_{i,t}} [\| G_i(\theta_t;B_{i,t}) \|^2 ] \leq \sigma_g^2.
\end{equation}
\textbf{Assumption 5:} (\textbf{Normalized Layer Selection Noise Bound}) There exist some $\xi^2 \in [0, 1)$ and any $t, i$, the normalized layer selection noise due to LNTK is bounded by:
\begin{equation}
\small
\frac{\| \theta_t - \hat{\theta}_{t,i}\|^2}{\| \theta_t \|^2} \leq \xi^2
\end{equation}
\noindent
where $\hat{\theta}_{t,i}$ denotes LNTK-based client-specific layers.

\noindent
\textbf{Theorem 2} (Impact of layer selection-based noise $\xi$ and variance of selection count $s_d^2$ on LNTK convergence):
\begin{equation}
\small
\begin{aligned}
&\min_{t \in [T]} \mathbb{E} \left[ \left\| \nabla F(\theta_t) \right\|_2^2 \right] \leq \frac{2}{(\eta - 3\gamma \eta^2) T} \Bigg[ F(\theta^0) - F(\theta^*) \\
 &+ \sum_{t=1}^{T} \frac{\eta \xi^2 \gamma^2 N s_d}{2} \left( 1 + 3 \eta \gamma \right) \mathbb{E} \left[ \left\| \sum_{t=1}^T \nabla_\ell F(\theta_t) \right\|^2 \right] \\
 &\quad + \frac{\gamma \eta^2 N T s_d}{2} \left(\eta \gamma \sigma_g^2 (1 + 3 \gamma) + 3 \sigma^2 s_d\right) \Bigg] \text{s.t.} \quad \theta^* = \arg \min_{\theta \in \mathbb{R}^p} F(\theta)
\end{aligned}
\end{equation}

\noindent
\textbf{Remark 2:} With $\eta \leq \text{min}\{\frac{1}{\sqrt{T}},\frac{1}{6\gamma}\}$, model converges to a neighborhood of a stationary point of FL with a small gap due to layer selection-based noise $\xi$ and variance of selection count $s_d$ over clients. This motivates us to jointly minimize the influence of $\xi$ and $s_d$ for better convergence. 
\subsection{Multi-objective Optimization}
Motivated by this, we refine the selected layers on server (see Fig. 6) to achieve two primary objectives (see Eq. 21): maximizing the cumulative client-specific importance scores based on LNTK and minimizing the variance of layer selection histogram (see Fig. 7) to encourage more balanced usage of each layer over clients. Let \( n_l \) represent the number of clients that select layer \( l \), with \( \bar{n} =\frac{1}{L} \sum_{l=1}^{L} n_l\) as the mean count of layer usage. (\textbf{See Suppl. \S A for more details.}) The joint optimization problem is formulated as:
\begin{equation}
\left\{
\begin{array}{ll}
\mathbf{\max} & \sum\limits_{i=1}^{\mathcal{N}} \sum\limits_{l=1}^L S_{i}^{l} \\
\mathbf{\min} & \frac{1}{L} \sum\limits_{l=1}^{L} \left( n_l - \bar{n} \right)^2
\end{array}
\right.
\text{s.t:}
  \sum_{l \in \mathcal{L}_i} m_{i}^{l} \leq L_{\text{i,max}}, \quad \forall i \in \mathcal{N}  
\end{equation}
The latter objective prevents over-reliance on specific layers even if they have high importance scores thereby increasing diversity in layer selection across clients. The constraint incorporates client-specific computational budget $L_{\text{i,max}}$.

\begin{figure}[t]
    \centering
\includegraphics[width=1.0\columnwidth]{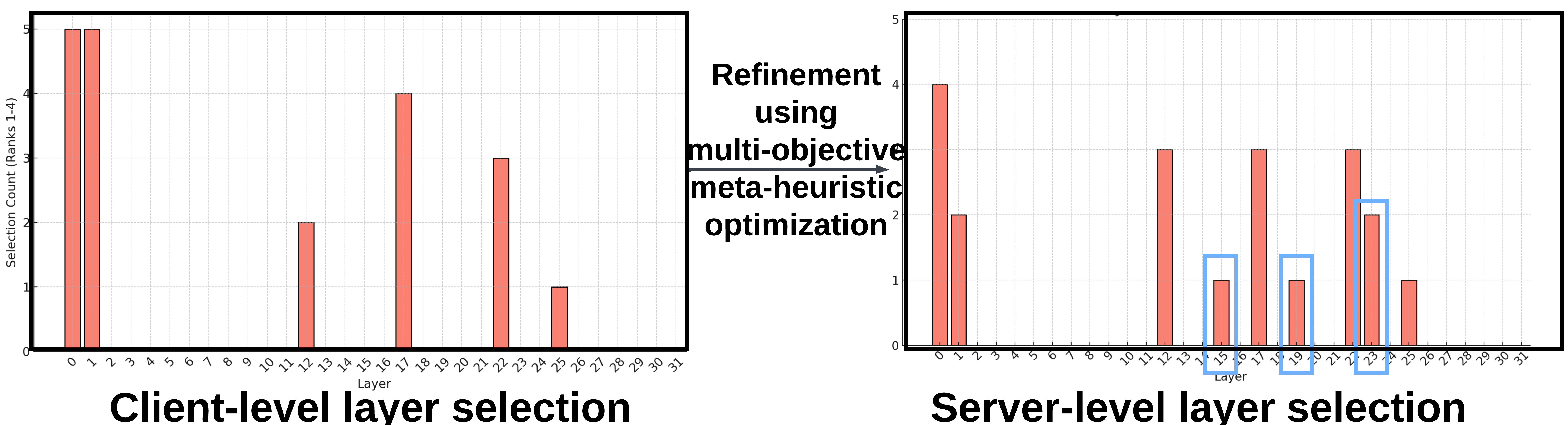}
    \caption{Layer selection histogram shows the impact of server-level optimization. It encourages more layers to participate (see blue-marked areas) by maximizing layer-selection diversity \textit{i.e.},  reducing the variance of the selection count (vertical axis).} 
\label{fig:11}
\end{figure}

\begin{figure*}[t]
    \centering
\includegraphics[width=2.0\columnwidth]{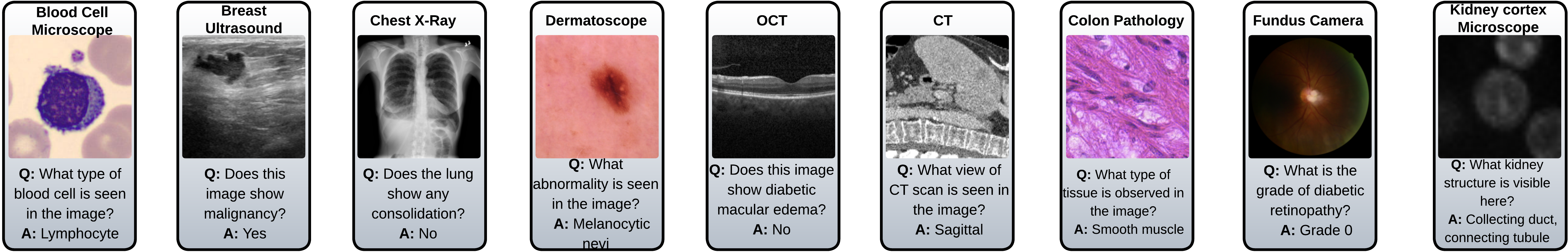}
    \caption{Sample VQA triplets of 9 modality-specific medical clients from Ultra-MedVQA \textbf{(\color{violet}Task 3\color{black})}} 
\label{fig:11}
\end{figure*}

Traditional optimization methods struggle to optimize these two conflicting objectives. Neural networks cannot be employed to optimize due to absence of data on server. Besides, the number of possible configurations is particularly high due to the large number of layers in foundation models. Since this problem involves multiple objectives, there is no "best" solution but rather a set of optimal trade-offs (the Pareto front). Meta-heuristic algorithms are well-suited to explore such complex, high-dimensional solution spaces in absence of training data by balancing exploration and exploitation.
To this end, we carefully select and investigate 5 meta-heuristic algorithms spanning all 3 algorithm categories: evolutionary, physics-based, and swarm-based. \textbf{See Suppl. \S A for detailed description and pseudo-code.}
\subsection{Meta-heuristic algorithms}
Each of the following algorithms aim to maximize client-specific importance while ensuring balanced layer utilization across clients based on their underlying principles:

\noindent
(1) \textbf{Non-Dominated Sorting Genetic Algorithm (NSGA):} We use NSGA \cite{deb2002fast} to iteratively evolve a population of layer selections. We initialize population based on client-specific importance scores while incorporating probability-based sampling for broader search space coverage. This guided randomness helps balance \textbf{exploration} (diversifying choices) and \textbf{exploitation} (prioritizing high-importance layers). Genetic operations like \textbf{crossover} and \textbf{mutation} generate new solutions, while \textbf{non-dominated sorting} and \textbf{crowding distance} ensure diversity on the Pareto Front.

\noindent
(2) \textbf{Artificial Bee Colony (ABC) \cite{bansal2013artificial}:} Each bee represents a potential layer selection for the clients. The optimization proceeds in three phases: \textbf{Employed bees} exploit local solutions, adjusting layer assignments based on importance scores and diversity, occasionally accepting worse solutions to escape local optima. \textbf{Onlooker bees} then select solutions based on a probability weighted by importance and diversity, while \textbf{Scout bees} abandon unproductive solutions, replacing them with randomly generated ones to promote exploration. Non-dominated solutions are stored in a Pareto archive, which guides refinement over iterations.

\noindent
(3) \textbf{Ant Colony Optimization (ACO) \cite{dorigo2006ant}:} Each ant represents a candidate layer selection, constructing paths influenced by \textbf{pheromone trails} (which reinforce successful layer choices from previous iterations) and \textbf{importance scores} (which guide ants toward layers that are likely beneficial based on client-specific requirements). The Pareto archive preserves non-dominated solutions. Pheromone trails are then updated, with evaporation to prevent stale paths and additional pheromone deposits on layers selected, thereby encouraging their selection in subsequent iterations.

\noindent
(4) \textbf{Simulated Annealing (SA):} SA \cite{van1987simulated} starts with a high-temperature, randomly initialized solution, gradually exploring optimal layer selections by cooling over iterations. It accepts new configurations if it offers higher importance or lower variance based on Eq. 21. If the new configuration is less optimal, it may still be accepted based on a probability that decreases with the temperature allowing SA to avoid being trapped in local optima. As the temperature cools, SA gradually refines the search, focusing on fine-tuning layer assignments that respect both client-specific importance and a balanced distribution of layer usage.

\noindent
(5) \textbf{Multi-Objective Particle Swarm Optimization (MOPSO) \cite{coello2002mopso}:}  
Each particle in the swarm represents a candidate layer assignment, initialized with client-specific importance scores to guide the selection of important layers. Randomized probability-based sampling is used to ensure diverse initial positions, promoting exploration of the solution space. Each particle updates its \textbf{velocity} and \textbf{position} by balancing three influences: \textbf{inertia} (maintaining its current layer selection), \textbf{a cognitive component} (best individual solution), and \textbf{a social component} (globally optimal solution). 

\begin{table*}[t]
    \centering
    \caption{Performance comparison on VLM adapter layer selection in terms of accuracy for Tasks 1-3 and F1-score for Tasks 4-6}
    \scalebox{0.67}{
    \begin{tabular}{l@{\hskip 0.05in}|c@{\hskip 0.05in}c@{\hskip 0.05in}c@{\hskip 0.05in}|c@{\hskip 0.05in}c@{\hskip 0.05in}c@{\hskip 0.05in}|c@{\hskip 0.05in}c@{\hskip 0.05in}c@{\hskip 0.05in}|c@{\hskip 0.05in}c@{\hskip 0.05in}c@{\hskip 0.05in}|c@{\hskip 0.05in}c@{\hskip 0.05in}c@{\hskip 0.05in}|c@{\hskip 0.05in}c@{\hskip 0.05in}c@{\hskip 0.05in}|c@{\hskip 0.05in}c@{\hskip 0.05in}c}
          \hline
        \textbf{Fine-tuning} & \multicolumn{3}{c|}{\textbf{\color{magenta}Task 1\color{black}}}& \multicolumn{3}{c|}{\textbf{\color{cyan}Task 2\color{black}}} & \multicolumn{3}{c|}{\textbf{\color{violet}Task 3\color{black}}} & \multicolumn{3}{c|}{\textbf{\color{black}\color{blue}Task 4\color{black}}} & \multicolumn{3}{c|}{\textbf{\color{red}Task 5\color{black}}} & \multicolumn{3}{c|}{\textbf{\color{brown}Task 6\color{black}}} & \multicolumn{3}{c}{\textbf{Mean Score}} \\
        
        \hline
        & \textbf{ViLT} & \textbf{LlAVA} & \textbf{BLIP} & \textbf{ViLT} & \textbf{LlAVA} & \textbf{BLIP} & \textbf{ViLT} & \textbf{LlAVA} & \textbf{BLIP} & \textbf{ViLT} & \textbf{LlAVA} & \textbf{BLIP} & \textbf{ViLT} & \textbf{LlAVA} & \textbf{BLIP} & \textbf{ViLT} & \textbf{LlAVA} & \textbf{BLIP} & \textbf{ViLT} & \textbf{LlAVA} & \textbf{BLIP} \\
        \hline
        All adapters & 43.04 & 40.02 & 43.36 & 82.46 & 79.70 & 79.03 & 78.55 & 75.19 & 76.40 & 63.05 & 76.89 & 71.05 & 65.68 & 77.93 & 76.30 & 85.83 & 88.02 & 89.26 & 69.77 & 72.96 & 72.57\\ \hline
        \textbf{} & \multicolumn{18}{c}{\textbf{Homogeneous resources across clients (L=4 for all clients)}}\\
        \hline
        FD \cite{wen2022federated} & 33.54 & 33.31 & 34.15 & 73.94 & 68.84 & 68.78 & 70.36 & 67.04 &  68.54& 52.82  & 67.78 & 60.21 & 56.20 & 67.99 & 66.53 & 76.90 & 79.05 & 80.36& 60.63 & 64.00 & 63.10\\ 
        Last \cite{ruckle2020adapterdrop} & 33.91 & 33.04 & 35.19 & 70.53 & 65.63 & 66.92 &68.32 & 65.03 & 66.59 & 54.40 & 65.94 & 58.45 & 57.77 & 66.05 & 64.20 & 77.62 & 79.51 & 79.96 & 60.42 & 62.53 & 61.88\\ 
        Magnitude \cite{han2015learning} & 31.60 & 32.40 & 30.26 & 70.32 & 67.35 & 67.67 & 69.18 &  68.73& 69.02 & 54.33 & 66.59 & 61.80 & 56.01 & 66.90 & 64.98 & 77.14 &80.26  & 80.44 & 59.76&63.71&62.36\\ 
        FishMask \cite{sung2021training} & 37.45 & {35.34} & 37.77 & 75.65 & {72.37} & 71.38 & 71.99 & 71.63 & 72.10 & 56.02 & 72.04 & 65.35 & 59.20 & 71.86 & 70.22 & 79.98 & 83.03 & 82.32& 63.38 & 67.71 & 66.52\\ 
        GradFlow \cite{lubanagradient} & 36.03 & {34.29} & 38.35 & 74.81 & 72.65 & 72.58 & 72.29 & 69.88 & 71.09 & 56.38 & 71.19 & 64.98 & 59.82 & 71.60 & 70.13 & 80.33 & 82.89 & 82.56& 63.28 & 67.08  & 66.61\\ 
        GraSP \cite{wang2020picking}& 35.46 & {34.90} & 38.88 & 75.03 & 72.03   & 71.74 & 72.03& 71.39 & 71.76 & 55.83 & 70.73 & 65.43 & 60.00 & 70.07 & 69.20 & 81.02 & 83.48 & 82.19& 63.23 & 67.10  & 66.53\\ 
        SNIP \cite{lee2018snip} & 31.26 & 32.73 & 35.36 & 73.96 & 69.40 & 70.15 & 72.16 & 67.04 & 68.76 & 54.21 & 66.07 & 60.39 & 58.09 & 67.24 & 66.92 & 78.20 & 80.34 & 81.13 & 61.31 & 63.80 &63.78\\ 
        RGN \cite{lee2022surgical}& 32.71 & 32.52 & 34.81 & 73.60 & 69.94 & 70.55 & 70.88 & 68.75  & 69.73 &56.40  & 67.25 & 61.32 & 57.54 & 68.64 & 68.39 & 76.79 & 78.05 & 80.09 & 61.32 & 64.19 & 64.15\\ 
        
        Synflow \cite{tanaka2020pruning}& 35.68 & 35.38 & 37.89 & 75.26 & 72.33 & 73.25 & 72.49 &71.36  & 71.63 & 56.11 & 71.67 & 64.28 & 59.27 & 70.17 & 71.32 & 80.35 & 79.09 & 81.90& 63.19 & 66.67 &66.71\\ 
        Fedselect \cite{tamirisa2024fedselect} & 36.80 & 34.48 & 38.88 & 74.29 & 70.49 &  71.63 & 71.29 & 70.52 & 70.06 & 55.83 & 70.86 & 65.08 & 58.53 & 71.33 & 72.02 & 79.39 & 83.22 & 83.00& 62.69 & 66.82 & 66.78\\
        SPT \cite{he2023sensitivity} & 35.59 & {34.40} & 38.53 & 75.50 & 72.16 & 71.32 & 72.58 &70.78  & 71.56 & 56.74 & 71.98 & 65.71 & 59.71 & 71.45 & 71.35 & 80.13 & 83.89 & 82.99 & 63.38 & 67.44  & 66.91\\
        LNTK (ours) & 39.74 & 36.80 & 40.43 & 77.54 & 74.64 & 74.49 & 74.01 & 72.68 & 73.93 & 58.07 & 73.06 &  67.80 & 61.50 &  73.00 & 73.43 & 82.04 & 85.79 & 84.84& 65.48 & 69.33  & 69.15\\ 
        \textbf{F\textsuperscript{3}OCUS} (ours) & \textbf{42.41} & \textbf{39.85} & \textbf{43.04} & \textbf{81.31} & \textbf{78.78} & \textbf{78.70}  & \textbf{77.86} & \textbf{75.01} & \textbf{76.45} & \textbf{62.51} & \textbf{76.70} & \textbf{70.35} & \textbf{64.62} & \textbf{77.26} & \textbf{76.14} & \textbf{85.28} & \textbf{87.53} & \textbf{88.30} & \textbf{69.00} & \textbf{72.52} & \textbf{72.16}\\ \hline
        \textbf{} & \multicolumn{18}{c}{\textbf{Heterogeneous resources across clients}}\\
        \hline
FD \cite{wen2022federated} & 32.65 & 33.54 & 34.04 & 73.74 & 69.36 & 68.16 & 70.27 & 66.73 & 67.81 & 52.18 & 67.51 & 60.48 & 55.56 & 67.71 & 66.16 & 76.85 & 78.57 & 79.54 & 60.21 & 63.90 & 62.70 \\ 
Last \cite{ruckle2020adapterdrop} & 33.65 & 33.80 & 35.30 & 70.90 & 65.81 & 66.97 & 68.22 & 65.92 & 66.31 & 54.89 & 66.35 & 58.43 & 57.91 & 65.47 & 64.75 & 77.86 & 78.80 & 80.12 & 60.57 & 62.69 & 61.98 \\ 
Magnitude \cite{han2015learning} & 32.04 & 32.54 & 30.81 & 69.84 & 67.00 & 67.53 & 68.54 & 68.48 & 68.57 & 53.83 & 66.19 & 61.46 & 56.12 & 66.30 & 65.11 & 77.68 & 80.48 & 80.78 & 59.68 & 63.50 & 62.38 \\ 
FishMask \cite{sung2021training} & 37.43 & 35.73 & 37.78 & 75.60 & 72.01 & 71.70 & 71.78 & 71.34 & 71.45 & 56.58 & 72.08 & 65.17 & 59.40 & 72.25 & 70.37 & 80.49 & 82.66 & 82.01 & 63.55 & 67.68 & 66.41 \\ 
GradFlow \cite{lubanagradient} & 35.17 & 34.86 & 38.43 & 75.41 & 73.16 & 71.99 & 72.45 & 70.40 & 71.20 & 56.33 & 70.91 & 65.17 & 59.44 & 71.48 & 70.19 & 80.66 & 83.39 & 82.96 & 63.24 & 67.37 & 66.66 \\ 
GraSP \cite{wang2020picking} & 36.11 & 35.20 & 38.68 & 74.99 & 72.38 & 70.99 & 71.54 & 71.54 & 71.98 & 55.50 & 71.12 & 65.24 & 59.50 & 69.63 & 69.17 & 81.85 & 82.63 & 81.48 & 63.25 & 67.08 & 66.26 \\ 
SNIP \cite{lee2018snip} & 30.53 & 33.39 & 36.24 & 73.96 & 68.70 & 70.33 & 72.00 & 66.63 & 68.03 & 54.02 & 66.76 & 60.09 & 58.82 & 67.15 & 66.90 & 78.01 & 79.66 & 81.14 & 61.22 & 63.72 & 63.79 \\ 
RGN \cite{lee2022surgical} & 31.82 & 32.64 & 35.14 & 73.74 & 70.12 & 70.52 & 70.49 & 69.04 & 69.13 & 56.02 & 67.68 & 61.42 & 58.22 & 69.21 & 68.50 & 77.04 & 78.52 & 79.75 & 61.22 & 64.54 & 64.08 \\ 
Synflow \cite{tanaka2020pruning} & 35.80 & 36.02 & 38.33 & 75.70 & 72.17 & 72.57 & 72.63 & 71.47 & 70.97 & 57.02 & 72.26 & 64.49 & 59.23 & 70.09 & 70.65 & 80.40 & 78.53 & 82.78 & 63.46 & 66.76 & 66.63 \\ 
Fedselect \cite{tamirisa2024fedselect} & 36.59 & 34.21 & 39.00 & 74.85 & 69.71 & 70.94 & 71.12 & 70.77 & 69.85 & 55.71 & 70.70 & 64.35 & 58.07 & 71.51 & 71.79 & 79.26 & 83.42 & 83.45 & 62.60 & 66.72 & 66.56 \\ 
SPT \cite{he2023sensitivity} & 35.35 & 34.72 & 38.58 & 75.34 & 72.84 & 71.82 & 72.92 & 70.87 & 72.10 & 56.23 & 72.65 & 65.38 & 59.38 & 72.05 & 71.20 & 80.41 & 83.82 & 82.44 & 63.27 & 67.82 & 66.92 \\ 
LNTK (ours) & 39.24 & 37.56 & 40.02 & 77.59 & 74.55 & 74.96 & 73.14 & 73.37 & 74.68 & 58.12 & 73.40 & 67.50 & 62.32 & 73.34 & 73.65 & 82.44 & 86.32 & 84.96 & 65.48 & 69.76 & 69.29 \\ 
\textbf{F\textsuperscript{3}OCUS} (ours) & \textbf{41.80} & \textbf{40.06} & \textbf{43.42} & \textbf{81.85} & \textbf{77.83} &\textbf{ 79.13} & \textbf{77.16} & \textbf{75.04} & \textbf{76.33} & \textbf{62.18} & \textbf{76.69} & \textbf{71.08} & \textbf{64.42} & \textbf{77.59} & \textbf{75.73} & \textbf{85.28} & \textbf{87.56} & \textbf{88.45} & \textbf{68.78} & \textbf{72.46} & \textbf{72.36} \\ 
        \hline
      
    \end{tabular}}
\end{table*}

\section{Experiments and Results}

\subsection{FL settings, Datasets and Tasks}
We evaluate our performance for fine-tuning selected (i) layers, and (ii) adapters \cite{houlsby2019parameter} with 4 VLMs of varying size and architecture, \textit{viz}., ViLT \cite{kim2021vilt}, ALBEF \cite{li2021align}, LlAVA-1.5 \cite{liu2024improved}, and BLIP-2 \cite{li2023blip}, for 3 FL task settings: (a) Visual Question Answering, (b) Image and Text-based Disease Classification, (c) Heterogeneous tasks combining (a), (b). 

\noindent
\textbf{(a) Visual Question Answering:} We consider \textbf{3} scenarios with data of varying sizes, class counts, and complexity: 
\noindent
\begin{itemize}
    \item[(i)] \textbf{\color{magenta}{Task 1 (with Domain gap)}\color{black} : Five-client setting} with SLAKE \cite{liu2021slake}, VQA-RAD \cite{lau2018dataset}, VQA-Med 2019 \cite{ben2019vqa}, VQA-Med 2020 \cite{BenAbacha2020VQAMed}, and VQA-Med 2021 \cite{ben2021overview}.
    
    \item[(ii)] \textbf{\color{cyan}{\color{cyan}{Task 2 (with Modality gap)}\color{black} }\color{black}: Modality specific 8-client setting} based on \cite{saha2024fedpia,Hu_2024_CVPR} with CT, Ultrasound, Dermatoscopy, Fundus, Histology, Microscopy, OCT, and X-Ray clients.

    \item[(iii)] \textbf{\color{violet}{{Task 3 (with Modality gap)}\color{black} }\color{black}: Modality specific 9-client setting} with our Ultra-MedVQA dataset \textbf{(see Fig. 8)}.
\end{itemize}
\noindent
\textbf{(b) Image and text-based multi-label disease classification:} We consider \textbf{2} FL settings \cite{saha2024examiningmodalityincongruitymultimodal} with Dirichlet coefficient $\gamma=0.5$ for Chest X-Ray and Radiology report-based multi-label disease detection (with 15 classes).
\begin{itemize}
\item [(i)] \textbf{\color{blue}{Task 4 (with label shift)}\color{black} :} \textbf{4 client-scenario} with Open-I 

\item [(ii)] \textbf{\color{red}{Task 5 (w/ label shift)}\color{black} :} \textbf{10 client scenario} with MIMIC. 
\end{itemize}


\begin{table}[t]
    \centering
    \caption{Comparison with other PEFTs on \textbf{\color{magenta}Task 1\color{black}} using \textbf{ALBEF}}
    \scalebox{0.6}{
    \begin{tabular}{l@{\hskip 0.05in}|c@{\hskip 0.05in}c@{\hskip 0.05in}c@{\hskip 0.05in}|c@{\hskip 0.05in}c@{\hskip 0.05in}c@{\hskip 0.05in}c@{\hskip 0.05in}c@{\hskip 0.05in}|c}
      
        \hline  
        \textbf{Method} & \textbf{MBit} & \textbf{GFLOP} & \textbf{Param(M)} & \color{magenta}\textbf{SLAKE} &  \color{magenta}\textbf{VM2019} & \color{magenta}\textbf{VM2020} & \color{magenta}\textbf{VM2021} & \color{magenta}\textbf{VR}   &   \color{magenta}\textbf{Overall} \\
        \hline 
        Full  & 3915.2 & 156.5 & 122.35 & 77.45 & 67.25 & 15.06 & 22.00 & 41.52 & 46.92  \\  
        Adap (all) & 28.8 & 85.2 & 0.90 & 72.82 & 64.45 & 11.56 & 21.00 & 38.35 & 43.17   \\  
        FedDAT & 86.1 & 86.3 &  2.69& 72.79 & 63.58 & 12.46 & 23.00 & 38.91 & 43.93 \\ \hline
        LN & 5.8 & 84.7 &  0.18 & 69.53 & 62.22 & 10.59 & 22.00 & 36.89 & 41.30 \\
        LoRA & 19.5 & 84.6 & 0.61 & 57.73 & 60.18 & 4.59 & 15.00 & 31.16 & 35.09 \\
        Bias & 3.5 & 84.7 & 0.11 & 68.25 & 55.61 & 11.17 & 17.00 & 35.47 & 39.54 \\
        PromptFL &19.2&85.0&0.60 & 65.63 &  57.56  & 4.78  & 15.00 & 40.26  & 36.65 \\
        \textbf{F\textsuperscript{3}OCUS} & 9.7 & 80.6 & 0.30 & 74.69 & 60.05 & 12.84 & 24.00 & 42.30 & 42.78 \\ 
        \hline
    \end{tabular}}
\end{table}

\noindent
\textbf{(c) Heterogeneous tasks:} We consider \textbf{\color{brown}{Task 6 (with task heterogeneity)}\color{black} } combining three Visual Question answering clients, \textit{viz}., SLAKE, VQA-RAD, VQA-Med 2019, and two disease-classification clients, \textit{viz}., Open-I and MIMIC.

\noindent
\textbf{Device Heterogeneity:} In Tab. 2 and 5, to simulate varying resource constraints among clients, we adjust the number of trainable layers across different tasks. For Tasks 1 and 6, 6 layers are finetuned for 2 clients, 4 layers for another 2 clients, and 2 layers for the remaining client. For Task 2, 6 layers are finetuned for 3 clients, 4 layers for 3 clients, and 2 layers for the last 2 clients. For Task 3, 2 layers are finetuned for 3 clients, 4 layers for another 3 clients and 6 layers for the last 3 clients. For Task 4, 6 layers are finetuned for 1 client, 4 layers for another client, and 2 layers for the remaining 2 clients. For Task 5, 6 layers are finetuned for 3 clients, 4 layers for 1 client, and 2 layers for the last client.

\noindent
\textbf{See Suppl. \S C for dataset and implementation details}
\subsection{Performance comparison with State-of-the-arts}
We compare \textbf{F\textsuperscript{3}OCUS} with \textbf{28 SOTA} methods: 

\noindent
(i) Tab. 3 shows comparison with \textbf{5 SOTA PEFT baselines}, \textit{viz.}, LayerNorm Tuning (LN) \cite{basu2023strong}, LoRA \cite{hu2022lora}, Bias Tuning \cite{tinyTL}, Prompt Tuning \cite{guo2023promptfl}, and FedDAT \cite{chen2024feddat} in terms of communication (MBits), Computation (GLOPs), total number of trainable parameters (Millions) and accuracy in each client. \textbf{F\textsuperscript{3}OCUS} is observed to outperform all PEFTs except adapters \cite{houlsby2019parameter} and FedDAT which finetune all adapters whereas \textbf{F\textsuperscript{3}OCUS} finetunes only selected 4 adapter layers in each client leading to reduced communication (9.7 MBits) and computational needs (80.6 GFLOPs).  

\noindent
(ii) In Tab. 6, \textbf{F\textsuperscript{3}OCUS} is seen to consistently outperform \textbf{12 SOTA Personalized FL baselines} \textit{viz.} perFedavg \cite{fallah2020personalized}, MetaFed \cite{chen2023metafed}, FedPAC \cite{xu2023personalized}, FedAS \cite{yang2024fedas}, FLUTE \cite{liu2024federated}, FedALA \cite{zhang2023fedala}, FedProto \cite{tan2022fedproto}, FedRod \cite{chen2021bridging}, FedAP \cite{lu2022personalized}, FedFomo \cite{zhang2020personalized}, FedRep \cite{collins2021exploiting}, and Fedper \cite{arivazhagan2019federated} for 5 tasks. 

\noindent
(iii) We adapt 7 \textbf{SOTA Pruning baselines} \textit{viz.} Federated Drop-out \cite{wen2022federated}, Magnitude \cite{han2015learning}, FishMask \cite{sung2021training}, GradFlow \cite{lubanagradient}, GraSP \cite{wang2020picking}, SNIP \cite{lee2018snip}, and Synflow \cite{tanaka2020pruning} in our context and compare with proposed method in Tabs. 2 and 5. LNTK surpasses the performance of the closest SOTA pruning method by \textbf{2.11$\%$} and \textbf{2.30$\%$} while \textbf{F\textsuperscript{3}OCUS} outperforms it by \textbf{5.35$\%$} and \textbf{5.33$\%$} respectively for homogeneous and heterogeneous device settings over all architectures. 

\noindent
(iv) We also compare with \textbf{4 SOTA Layer Selection baselines} \textit{viz.} Adapter-drop (denoted as 'last') \cite{ruckle2020adapterdrop}, RGN \cite{lee2022surgical}, Fedselect \cite{tamirisa2024fedselect}, and SPT \cite{he2023sensitivity} (see Tabs. 2, 5). From Table 2, we observe that LNTK outperforms the closest SOTA method by \textbf{2.08$\%$} and \textbf{2.17$\%$} for homogeneous and heterogeneous resource settings respectively. \textbf{F\textsuperscript{3}OCUS} further improves performance over LNTK by\textbf{ 3.24$\%$} and \textbf{3.03$\%$}.

\begin{figure}[t]
    \centering
\includegraphics[width=1.0\columnwidth]{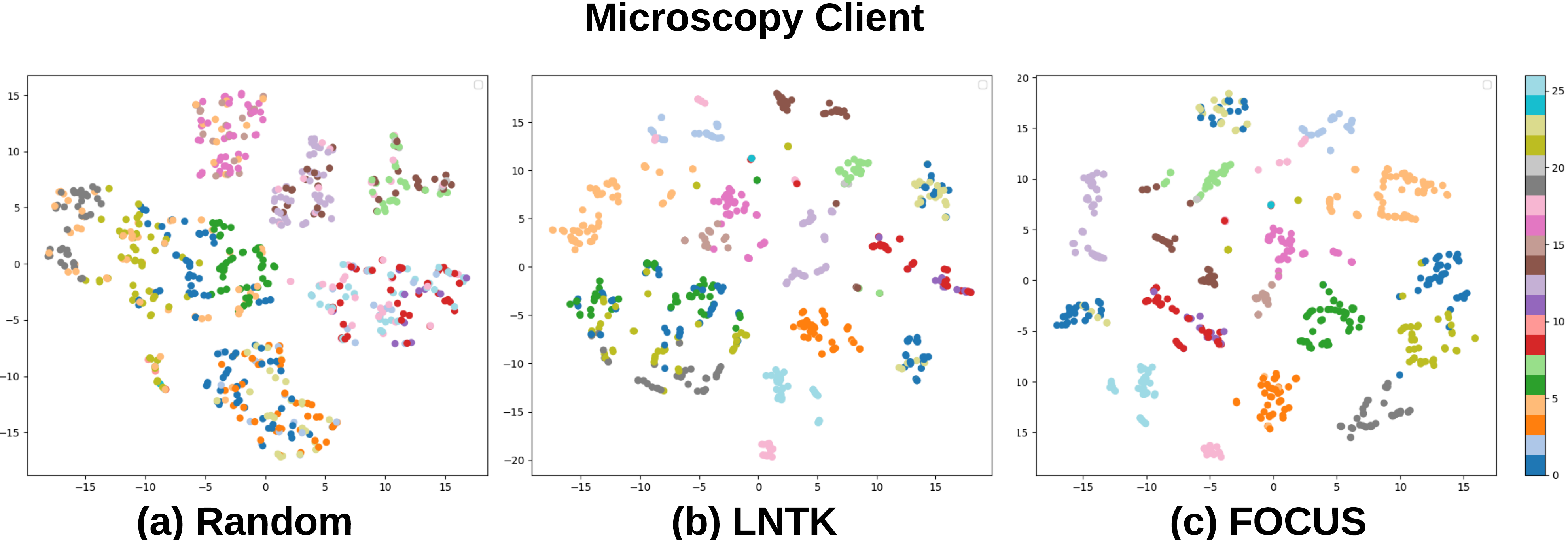}
    \caption{t-SNE feature visualization for Client 6 of Task 2 (with 26 classes) shows greater separability for \textbf{F\textsuperscript{3}OCUS}.}
\label{fig:11}
\end{figure}

\begin{table}[t]
    \centering
    \caption{Comparison of meta-heuristic methods (\color{cyan}{Task 2} \color{black})}
    \scalebox{0.67}{
    \begin{tabular}{l@{\hskip 0.05in}|c@{\hskip 0.05in}c@{\hskip 0.05in}c@{\hskip 0.05in}c@{\hskip 0.05in}c@{\hskip 0.05in}c@{\hskip 0.05in}c@{\hskip 0.05in}c@{\hskip 0.05in}|c}
        \hline
        \textbf{Finetune}  & \color{cyan}\textbf{C1} & \color{cyan}\textbf{C2} & \color{cyan}\textbf{C3} & \color{cyan}\textbf{C4} & \color{cyan}\textbf{C5} & \color{cyan}\textbf{C6} & \color{cyan}\textbf{C7} & \color{cyan}\textbf{C8} & \color{cyan}\textbf{Overall} \\
        \ & {(CT)} & {(US)} & {(OCT)} & {(Fundus)} & {(Micro.)} & {(Hist.)} & {(Derma.)} & {(XRay)} &  \\
        \hline
  NSGA &      88.67 & \textbf{79.41} & 78.29 & 83.48 & 70.68 & 88.49 & \textbf{67.69} & 73.93 & \textbf{78.78} \\
ABC & \textbf{91.33} & 74.51 & 77.71 & 85.71 & \textbf{72.73} & 88.70  & 62.31 & \textbf{76.55} & 78.70  \\
ACO & 86.67 & 78.10  & \textbf{78.86} & 83.48 & 70.68 & 88.08 & 66.92 & 76.00  & 78.60  \\
SA & 88.67 & 72.59 & 70.86 & \textbf{85.93} & 71.09 & \textbf{89.45} & 66.62 & 76.48 & 77.71 \\
MOPSO & 86.18 & 77.31 & 78.83 & 84.53 & 71.18 & 87.58 & 65.49 & 75.97 & 78.38 \\ \hline
    \end{tabular}}
\end{table}

\subsection{Other Experimental Results} 
We plot the loss curves of LNTK and \textbf{F\textsuperscript{3}OCUS} and compare them with several pruning methods in \textbf{Fig. 2}, which demonstrates the impact of our server-level optimization on faster convergence. We also visualize the principal eigenvalue magnitudes of all 32 layers of LLaVA across different clients at the beginning of a round for Task 1 in \textbf{Fig. 3} and show the layer ranks over all rounds in \textbf{Fig. 5}. Consequently, the effect of server-level optimization on overall layer selections is shown in \textbf{Fig. 6} using relative layer ranks and in \textbf{Fig. 7} using the layer selection histogram. 
\textbf{Table 4} shows the comparable performance of different meta-heuristic algorithms for all clients in Task 2. For the 'microscopy' client in the same task, we present t-SNE feature visualizations in \textbf{Fig. 9} for (a) Federated Dropout (labeled random), (b) LNTK, and (c) \textbf{F\textsuperscript{3}OCUS}, highlighting the striking improvement in generating distinctive feature representations. 
\textbf{See Suppl. \S D for more results.}
\section{Conclusion}
We presented \textbf{F\textsuperscript{3}OCUS}, a novel and theoretically grounded approach for federated fine-tuning of Vision-Language foundation models in client-specific resource-constrained settings. \textbf{F\textsuperscript{3}OCUS} effectively balances individual client requirements with collective layer diversity thereby improving model convergence and performance accuracy. Our server-level multi-objective meta-heuristic optimization scheme does not require any data on server and can be easily combined with any existing layer selection or pruning algorithms. Additionally, we released \textbf{Ultra-MedVQA}, the largest medical VQA dataset, covering 12 anatomies, for supporting further VLM research. Experimental results demonstrate that \textbf{F\textsuperscript{3}OCUS} achieves superior performance across multiple vision-language tasks and models, providing a practical solution for fine-tuning large foundation models in decentralized environments.
\begin{table}[t]
    \centering
    \caption{Performance comparison on VLM layer selection with heterogeneous resources across clients}
    \scalebox{0.55}{
    \begin{tabular}{l@{\hskip 0.05in}|c@{\hskip 0.05in}c@{\hskip 0.05in}|c@{\hskip 0.05in}c@{\hskip 0.05in}|c@{\hskip 0.05in}c@{\hskip 0.05in}|c@{\hskip 0.05in}c@{\hskip 0.05in}|c@{\hskip 0.05in}c@{\hskip 0.05in}|c@{\hskip 0.05in}c}
        \hline
        \textbf{Fine-tuning} & \multicolumn{2}{c|}{\textbf{\color{magenta}Task 1\color{black}}} & \multicolumn{2}{c|}{\textbf{\color{cyan}Task 2\color{black}}} & \multicolumn{2}{c|}{\textbf{\color{violet}Task 3\color{black}}} & \multicolumn{2}{c|}{\textbf{\color{blue}Task 4\color{black}}} & \multicolumn{2}{c|}{\textbf{\color{red}Task 5\color{black}}} & \multicolumn{2}{c}{\textbf{\color{brown}Task 6\color{black}}} \\
        
        \hline
        & \textbf{ViLT} & \textbf{BLIP} & \textbf{ViLT} & \textbf{BLIP} & \textbf{ViLT} & \textbf{BLIP} & \textbf{ViLT} & \textbf{BLIP} & \textbf{ViLT} & \textbf{BLIP} & \textbf{ViLT} & \textbf{BLIP} \\
        \hline

FD \cite{wen2022federated} &   31.91 & 33.07 & 75.75 & 70.03 & 72.84 & 70.13 & 50.25 & 58.12 & 56.80 & 68.10 & 77.65 & 82.70  \\ 
Last \cite{ruckle2020adapterdrop} & 32.75 & 34.31 & 76.65 & 68.35 & 72.63 & 68.56 & 53.77 & 56.46 & 59.07 & 66.25 & 80.75 & 81.70  \\ 
Magnitude \cite{han2015learning} & 31.35 & 30.96 & 73.26 & 69.18 & 70.88 & 71.02 & 52.17 & 58.46 & 58.25 & 67.04 & 77.51 & 82.27  \\
FishMask \cite{sung2021training} & 36.35 & 36.82 & 77.71 & 73.56 & 74.60 & 73.61 & 54.35 & 63.08 & 61.10 & 73.39 & 81.90 & 83.09  \\ 
GradFlow \cite{lubanagradient} & 36.20 & 37.29 & 77.09 & 74.06 & 74.82 & 73.90 & 54.18 & 62.58 & 60.70 & 71.76 & 81.22 & 82.24\\
GraSP \cite{wang2020picking}&  35.36 & 37.87 & 76.90 & 73.00 & 73.80 & 74.83 & 53.44 & 62.68 & 60.91 & 70.85 & 82.03 & 82.54 \\ 
SNIP \cite{lee2018snip} & 29.65 & 35.23 & 75.59 & 73.88 & 74.37 & 72.14 & 51.91 & 58.04 & 62.04 & 68.85 & 79.25 & 82.38  \\ 
RGN \cite{lee2022surgical}&  32.89 & 36.21 & 75.34 & 72.35 & 72.76 & 71.59 & 53.80 & 62.16 & 59.58 & 70.23 & 77.95 & 81.14  \\
Synflow \cite{tanaka2020pruning}& 34.99 & 37.39 & 77.95 & 74.05 & 74.81 & 73.79 & 55.09 & 62.08 & 60.97 & 72.10 & 81.30 & 83.65  \\ 
Fedselect \cite{tamirisa2024fedselect} & 35.79 & 38.00 & 76.66 & 72.80 & 73.45 & 72.60 & 53.55 & 61.87 & 59.34 & 71.63 & 80.29 & 84.10  \\ 
SPT \cite{he2023sensitivity} & 34.33 & 37.78 & 77.03 & 73.66 & 75.18 & 74.80 & 54.18 & 63.03 & 60.60 & 72.68 & 81.53 & 83.97  \\ 
LNTK & 37.32 & 39.22 & 79.15 & 76.47 & 75.44 & 76.99 & 55.81 & 65.40 & 64.12 & 75.26 & 83.66 & 86.20  \\ 
\textbf{F\textsuperscript{3}OCUS} & \textbf{40.85} & \textbf{42.37} & \textbf{84.25} & \textbf{80.46} &\textbf{79.65} & \textbf{78.95} & \textbf{60.29} & \textbf{69.79} & \textbf{65.86} & \textbf{78.94} & \textbf{86.40} & \textbf{89.69}  \\ 

        \hline
    \end{tabular}}
\end{table}

\begin{table}
\centering
\caption{Comparison with SOTA personalized FL (L=4).  pF: perFedavg, MF: MetaFed, FP: FedPAC, FAS: FedAS, FT: FLUTE, FAL: FedALA, FPr: FedProto, FR: FedRod, FA: FedAP, FF: FedFomo, FR: FedRep, Fpe: Fedper}
\scalebox{0.65}{
\begin{tabular}{c@{\hskip 0.01in}|c|@{\hskip 0.01in}c|@{\hskip 0.01in}c|@{\hskip 0.01in}c|@{\hskip 0.01in}c|@{\hskip 0.01in}c|@{\hskip 0.01in}c|@{\hskip 0.01in}c|@{\hskip 0.01in}c|@{\hskip 0.01in}c|@{\hskip 0.01in}c|@{\hskip 0.01in}c|@{\hskip 0.01in}c@{\hskip 0.01in}}
\hline
\textbf{Task} & \textbf{pF} & \textbf{MF} & \textbf{FP} & \textbf{FAS} & \textbf{FT} & \textbf{FAL} & \textbf{FPr} & \textbf{FR} & \textbf{FA} & \textbf{FF} & \textbf{FRp} & \textbf{Fpe} & \textbf{\textbf{F\textsuperscript{3}OCUS}} \\
\hline
\color{magenta}1 & 33.9 & 35.4 & 36.2 & 36.5 & 30.8 & 35.5 & 35.5 & 36.2 & 36.8 & 35.4 & 35.3 & 36.1  & \textbf{42.4} \\ 
\color{cyan} 2 & 71.6 & 75.5 & 76.3 & 76.5 & 62.3 & 74.6 & 75.0 & 77.7 & 77.4 & 75.1 & 75.5 & 75.9& \textbf{81.3} \\ 
\color{blue}        4 & 54.6 & 57.3 & 57.5 & 58.8 & 44.8 & 57.2 & 57.0 & 59.4 & 59.5 & 57.1 & 58.4 & 58.1 & \textbf{62.5}\\ 
\color{red}        5 & 57.6 & 59.2 & 58.7 & 59.2 & 47.9 & 57.4 & 57.4 & 58.8 & 60.4 & 58.3 & 57.0 & 57.6  & \textbf{64.6} \\ 
\color{brown}        6 & 73.3 & 76.3 & 78.2 & 78.5 & 63.2 & 75.9 & 77.0 & 76.1 & 76.3 & 75.7 & 76.6 & 76.8 & \textbf{85.3} \\ \hline

\hline
\end{tabular}}
\end{table}

\section*{Acknowledgment}
This work was supported in part by the UK EPSRC (Engineering and Physical Research Council) Programme Grant EP/T028572/1 (VisualAI), a  UK EPSRC Doctoral Training Partnership award, the UKRI grant EP/X040186/1 (Turing AI Fellowship), and the InnoHK-funded Hong Kong Centre for Cerebro-cardiovascular Health Engineering (COCHE) Project 2.1 (Cardiovascular risks in early life and fetal echocardiography). FW is supported by the EPSRC Centre for Doctoral Training in Health Data Science (EP/S02428X/1), by the Anglo-Austrian Society, and by an Oxford-Reuben scholarship. D.A.C. is supported by the Pandemic Sciences Institute at the University of Oxford, the National Institute for Health Research (NIHR) Oxford Biomedical Research Center (BRC), an NIHR Research Professorship, a Royal Academy of Engineering Research Chair, the Wellcome Trust, the UKRI, and the InnoHK Hong Kong Center for Center for Cerebro-cardiovascular Engineering (COCHE).
{
    \small
    \bibliographystyle{ieeenat_fullname}
    \bibliography{main}

\begin{thebibliography}{94}
\providecommand{\natexlab}[1]{#1}
\providecommand{\url}[1]{\texttt{#1}}
\expandafter\ifx\csname urlstyle\endcsname\relax
  \providecommand{\doi}[1]{doi: #1}\else
  \providecommand{\doi}{doi: \begingroup \urlstyle{rm}\Url}\fi

\bibitem[Abacha et~al.()Abacha, Datla, Hasan, Demner-Fushman, and M{\"u}ller]{BenAbacha2020VQAMed}
Abeed S.~Ben Abacha, Vivek~V. Datla, Sadid~A. Hasan, Dina Demner-Fushman, and Henning M{\"u}ller.
\newblock Overview of the vqa-med task at imageclef 2020: Visual question answering and generation in the medical domain.
\newblock In \emph{CLEF 2020 Working Notes}.

\bibitem[Acar et~al.(2021)Acar, Zhao, Navarro, Mattina, Whatmough, and Saligrama]{acar2021federated}
Durmus Alp~Emre Acar, Yue Zhao, Ramon~Matas Navarro, Matthew Mattina, Paul~N Whatmough, and Venkatesh Saligrama.
\newblock Federated learning based on dynamic regularization.
\newblock \emph{arXiv preprint arXiv:2111.04263}, 2021.

\bibitem[Arivazhagan et~al.(2019)Arivazhagan, Aggarwal, Singh, and Choudhary]{arivazhagan2019federated}
Manoj~Ghuhan Arivazhagan, Vinay Aggarwal, Aaditya~Kumar Singh, and Sunav Choudhary.
\newblock Federated learning with personalization layers.
\newblock \emph{arXiv preprint arXiv:1912.00818}, 2019.

\bibitem[Bansal et~al.(2013)Bansal, Sharma, and Jadon]{bansal2013artificial}
Jagdish~Chand Bansal, Harish Sharma, and Shimpi~Singh Jadon.
\newblock Artificial bee colony algorithm: a survey.
\newblock \emph{International Journal of Advanced Intelligence Paradigms}, 5\penalty0 (1-2):\penalty0 123--159, 2013.

\bibitem[Basu et~al.()Basu, Massiceti, Hu, and Feizi]{basu2023strong}
Samyadeep Basu, Daniela Massiceti, Shell~Xu Hu, and Soheil Feizi.
\newblock Strong baselines for parameter efficient few-shot fine-tuning.
\newblock \emph{arXiv preprint arXiv:2304.01917}.

\bibitem[Ben~Abacha et~al.(2019)Ben~Abacha, Hasan, Datla, Demner-Fushman, and M{\"u}ller]{ben2019vqa}
Asma Ben~Abacha, Sadid~A Hasan, Vivek~V Datla, Dina Demner-Fushman, and Henning M{\"u}ller.
\newblock Vqa-med: Overview of the medical visual question answering task at imageclef 2019.
\newblock In \emph{Proceedings of CLEF (Conference and Labs of the Evaluation Forum) 2019 Working Notes}, 2019.

\bibitem[Ben~Abacha et~al.(2021)Ben~Abacha, Sarrouti, Demner-Fushman, Hasan, and M{\"u}ller]{ben2021overview}
Asma Ben~Abacha, Mourad Sarrouti, Dina Demner-Fushman, Sadid~A Hasan, and Henning M{\"u}ller.
\newblock Overview of the vqa-med task at imageclef 2021: Visual question answering and generation in the medical domain.
\newblock In \emph{Proceedings of the CLEF 2021 Conference and Labs of the Evaluation Forum-working notes}, 2021.

\bibitem[Ben~Zaken et~al.(2022)Ben~Zaken, Goldberg, and Ravfogel]{ben-zaken-etal-2022-bitfit}
Elad Ben~Zaken, Yoav Goldberg, and Shauli Ravfogel.
\newblock {B}it{F}it: Simple parameter-efficient fine-tuning for transformer-based masked language-models.
\newblock In \emph{Proceedings of the Association for Computational Linguistics (Volume 2: Short Papers)}, pages 1--9, Dublin, Ireland, 2022.

\bibitem[Bordes et~al.(2024)Bordes, Pang, Ajay, Li, Bardes, Petryk, Mañas, Lin, Mahmoud, Jayaraman, Ibrahim, Hall, Xiong, Lebensold, Ross, Jayakumar, Guo, Bouchacourt, Al-Tahan, Padthe, Sharma, Xu, Tan, Richards, Lavoie, Astolfi, Hemmat, Chen, Tirumala, Assouel, Moayeri, Talattof, Chaudhuri, Liu, Chen, Garrido, Ullrich, Agrawal, Saenko, Celikyilmaz, and Chandra]{bordes2024introductionvisionlanguagemodeling}
Florian Bordes, Richard~Yuanzhe Pang, Anurag Ajay, Alexander~C. Li, Adrien Bardes, Suzanne Petryk, Oscar Mañas, Zhiqiu Lin, Anas Mahmoud, Bargav Jayaraman, Mark Ibrahim, Melissa Hall, Yunyang Xiong, Jonathan Lebensold, Candace Ross, Srihari Jayakumar, Chuan Guo, Diane Bouchacourt, Haider Al-Tahan, Karthik Padthe, Vasu Sharma, Hu Xu, Xiaoqing~Ellen Tan, Megan Richards, Samuel Lavoie, Pietro Astolfi, Reyhane~Askari Hemmat, Jun Chen, Kushal Tirumala, Rim Assouel, Mazda Moayeri, Arjang Talattof, Kamalika Chaudhuri, Zechun Liu, Xilun Chen, Quentin Garrido, Karen Ullrich, Aishwarya Agrawal, Kate Saenko, Asli Celikyilmaz, and Vikas Chandra.
\newblock An introduction to vision-language modeling, 2024.

\bibitem[Bowman(2023)]{bowman2023brief}
Benjamin Bowman.
\newblock A brief introduction to the neural tangent kernel.
\newblock 2023.

\bibitem[Cai et~al.(2020)Cai, Gan, Zhu, and Han]{tinyTL}
Han Cai, Chuang Gan, Ligeng Zhu, and Song Han.
\newblock Tinytl: Reduce memory, not parameters for efficient on-device learning.
\newblock In \emph{Advances in Neural Information Processing Systems}, pages 11285--11297, 2020.

\bibitem[Cao et~al.(2019)Cao, Fang, Wu, Zhou, and Gu]{cao2019towards}
Yuan Cao, Zhiying Fang, Yue Wu, Ding-Xuan Zhou, and Quanquan Gu.
\newblock Towards understanding the spectral bias of deep learning.
\newblock \emph{arXiv preprint arXiv:1912.01198}, 2019.

\bibitem[Chen et~al.(2023{\natexlab{a}})Chen, Yao, Gao, Ding, and Li]{chen2023efficient}
Daoyuan Chen, Liuyi Yao, Dawei Gao, Bolin Ding, and Yaliang Li.
\newblock Efficient personalized federated learning via sparse model-adaptation.
\newblock In \emph{International Conference on Machine Learning}, pages 5234--5256. PMLR, 2023{\natexlab{a}}.

\bibitem[Chen et~al.(2024)Chen, Zhang, Krompass, Gu, and Tresp]{chen2024feddat}
Haokun Chen, Yao Zhang, Denis Krompass, Jindong Gu, and Volker Tresp.
\newblock Feddat: An approach for foundation model finetuning in multi-modal heterogeneous federated learning.
\newblock In \emph{Proceedings of the AAAI Conference on Artificial Intelligence}, pages 11285--11293, 2024.

\bibitem[Chen and Chao(2021)]{chen2021bridging}
Hong-You Chen and Wei-Lun Chao.
\newblock On bridging generic and personalized federated learning for image classification.
\newblock \emph{arXiv preprint arXiv:2107.00778}, 2021.

\bibitem[Chen et~al.(2022)Chen, Xu, Guo, Wang, Zhang, and Wang]{chen2022fedtune}
Jinyu Chen, Wenchao Xu, Song Guo, Junxiao Wang, Jie Zhang, and Haozhao Wang.
\newblock Fedtune: A deep dive into efficient federated fine-tuning with pre-trained transformers.
\newblock \emph{arXiv preprint arXiv:2211.08025}, 2022.

\bibitem[Chen et~al.(2023{\natexlab{b}})Chen, Lu, Qin, Wang, and Xie]{chen2023metafed}
Yiqiang Chen, Wang Lu, Xin Qin, Jindong Wang, and Xing Xie.
\newblock Metafed: Federated learning among federations with cyclic knowledge distillation for personalized healthcare.
\newblock \emph{IEEE Transactions on Neural Networks and Learning Systems}, 2023{\natexlab{b}}.

\bibitem[Coello and Lechuga(2002)]{coello2002mopso}
CA~Coello Coello and Maximino~Salazar Lechuga.
\newblock Mopso: A proposal for multiple objective particle swarm optimization.
\newblock In \emph{Proceedings of the 2002 Congress on Evolutionary Computation. CEC'02 (Cat. No. 02TH8600)}, pages 1051--1056. IEEE, 2002.

\bibitem[Collins et~al.(2021)Collins, Hassani, Mokhtari, and Shakkottai]{collins2021exploiting}
Liam Collins, Hamed Hassani, Aryan Mokhtari, and Sanjay Shakkottai.
\newblock Exploiting shared representations for personalized federated learning.
\newblock In \emph{International conference on machine learning}, pages 2089--2099. PMLR, 2021.

\bibitem[Deb et~al.(2002)Deb, Pratap, Agarwal, and Meyarivan]{deb2002fast}
Kalyanmoy Deb, Amrit Pratap, Sameer Agarwal, and TAMT Meyarivan.
\newblock A fast and elitist multiobjective genetic algorithm: Nsga-ii.
\newblock \emph{IEEE transactions on evolutionary computation}, 6\penalty0 (2):\penalty0 182--197, 2002.

\bibitem[Dorigo et~al.(2006)Dorigo, Birattari, and Stutzle]{dorigo2006ant}
Marco Dorigo, Mauro Birattari, and Thomas Stutzle.
\newblock Ant colony optimization.
\newblock \emph{IEEE computational intelligence magazine}, 1\penalty0 (4):\penalty0 28--39, 2006.

\bibitem[Dun et~al.(2022)Dun, Wolfe, Jermaine, and Kyrillidis]{dun2022resist}
Chen Dun, Cameron~R Wolfe, Christopher~M Jermaine, and Anastasios Kyrillidis.
\newblock Resist: Layer-wise decomposition of resnets for distributed training.
\newblock In \emph{Uncertainty in Artificial Intelligence}, pages 610--620. PMLR, 2022.

\bibitem[Fallah et~al.(2020)Fallah, Mokhtari, and Ozdaglar]{fallah2020personalized}
Alireza Fallah, Aryan Mokhtari, and Asuman Ozdaglar.
\newblock Personalized federated learning with theoretical guarantees: A model-agnostic meta-learning approach.
\newblock \emph{Advances in neural information processing systems}, 33:\penalty0 3557--3568, 2020.

\bibitem[Frankle et~al.(2021)Frankle, Schwab, and Morcos]{frankle2021trainingbatchnormbatchnormexpressive}
Jonathan Frankle, David~J. Schwab, and Ari~S. Morcos.
\newblock Training batchnorm and only batchnorm: On the expressive power of random features in cnns, 2021.

\bibitem[Guo et~al.(2023)Guo, Guo, Wang, Tang, and Xu]{guo2023promptfl}
Tao Guo, Song Guo, Junxiao Wang, Xueyang Tang, and Wenchao Xu.
\newblock Promptfl: Let federated participants cooperatively learn prompts instead of models-federated learning in age of foundation model.
\newblock \emph{IEEE Transactions on Mobile Computing}, 2023.

\bibitem[Han et~al.(2015)Han, Pool, Tran, and Dally]{han2015learning}
Song Han, Jeff Pool, John Tran, and William Dally.
\newblock Learning both weights and connections for efficient neural network.
\newblock \emph{Advances in neural information processing systems}, 28, 2015.

\bibitem[He et~al.(2023)He, Cai, Zhang, Tao, and Zhuang]{he2023sensitivity}
Haoyu He, Jianfei Cai, Jing Zhang, Dacheng Tao, and Bohan Zhuang.
\newblock Sensitivity-aware visual parameter-efficient fine-tuning.
\newblock In \emph{Proceedings of the IEEE/CVF International Conference on Computer Vision}, pages 11825--11835, 2023.

\bibitem[He et~al.(2020)He, Zhang, Mou, Xing, and Xie]{he2020pathvqa}
Xuehai He, Yichen Zhang, Luntian Mou, Eric Xing, and Pengtao Xie.
\newblock Pathvqa: 30000+ questions for medical visual question answering.
\newblock \emph{arXiv preprint arXiv:2003.10286}, 2020.

\bibitem[Hernandez-Cruz et~al.(2024)Hernandez-Cruz, Saha, Sarker, and Noble]{hernandez2024review}
Netzahualcoyotl Hernandez-Cruz, Pramit Saha, Md~Mostafa~Kamal Sarker, and J~Alison Noble.
\newblock Review of federated learning and machine learning-based methods for medical image analysis.
\newblock \emph{Big Data and Cognitive Computing}, 8\penalty0 (9):\penalty0 99, 2024.

\bibitem[Hilmkil et~al.(2021)Hilmkil, Callh, Barbieri, S{\"u}tfeld, Zec, and Mogren]{hilmkil2021scaling}
Agrin Hilmkil, Sebastian Callh, Matteo Barbieri, Leon~Ren{\'e} S{\"u}tfeld, Edvin~Listo Zec, and Olof Mogren.
\newblock Scaling federated learning for fine-tuning of large language models.
\newblock In \emph{International Conference on Applications of Natural Language to Information Systems}, pages 15--23. Springer, 2021.

\bibitem[Houlsby et~al.(2019)Houlsby, Giurgiu, Jastrzebski, Morrone, De~Laroussilhe, Gesmundo, Attariyan, and Gelly]{houlsby2019parameter}
Neil Houlsby, Andrei Giurgiu, Stanislaw Jastrzebski, Bruna Morrone, Quentin De~Laroussilhe, Andrea Gesmundo, Mona Attariyan, and Sylvain Gelly.
\newblock Parameter-efficient transfer learning for nlp.
\newblock In \emph{International conference on machine learning}, pages 2790--2799. PMLR, 2019.

\bibitem[Hu et~al.(2022)Hu, yelong shen, Wallis, Allen-Zhu, Li, Wang, Wang, and Chen]{hu2022lora}
Edward~J Hu, yelong shen, Phillip Wallis, Zeyuan Allen-Zhu, Yuanzhi Li, Shean Wang, Lu Wang, and Weizhu Chen.
\newblock Lo{RA}: Low-rank adaptation of large language models.
\newblock In \emph{International Conference on Learning Representations}, 2022.

\bibitem[Hu et~al.(2024)Hu, Li, Lu, Shao, He, Qiao, and Luo]{Hu_2024_CVPR}
Yutao Hu, Tianbin Li, Quanfeng Lu, Wenqi Shao, Junjun He, Yu Qiao, and Ping Luo.
\newblock Omnimedvqa: A new large-scale comprehensive evaluation benchmark for medical lvlm.
\newblock In \emph{Proceedings of the IEEE/CVF Conference on Computer Vision and Pattern Recognition (CVPR)}, pages 22170--22183, 2024.

\bibitem[Jacot et~al.(2018)Jacot, Gabriel, and Hongler]{jacot2018neural}
Arthur Jacot, Franck Gabriel, and Cl{\'e}ment Hongler.
\newblock Neural tangent kernel: Convergence and generalization in neural networks.
\newblock \emph{Advances in neural information processing systems}, 31, 2018.

\bibitem[Jia et~al.(2022)Jia, Tang, Chen, Cardie, Belongie, Hariharan, and Lim]{VPT_jia}
Menglin Jia, Luming Tang, Bor-Chun Chen, Claire Cardie, Serge Belongie, Bharath Hariharan, and Ser-Nam Lim.
\newblock Visual prompt tuning.
\newblock In \emph{Computer Vision – ECCV 2022: 17th European Conference, Tel Aviv, Israel, October 23–27, 2022, Proceedings, Part XXXIII}, page 709–727. Springer-Verlag, 2022.

\bibitem[Kaplun et~al.(2023)Kaplun, Gurevich, Swisa, David, Shalev-Shwartz, and Malach]{kaplun2023less}
Gal Kaplun, Andrey Gurevich, Tal Swisa, Mazor David, Shai Shalev-Shwartz, and Eran Malach.
\newblock Less is more: Selective layer finetuning with subtuning.
\newblock \emph{arXiv preprint arXiv:2302.06354}, 2023.

\bibitem[Karimireddy et~al.(2020)Karimireddy, Kale, Mohri, Reddi, Stich, and Suresh]{karimireddy2020scaffold}
Sai~Praneeth Karimireddy, Satyen Kale, Mehryar Mohri, Sashank Reddi, Sebastian Stich, and Ananda~Theertha Suresh.
\newblock Scaffold: Stochastic controlled averaging for federated learning.
\newblock In \emph{International conference on machine learning}. PMLR, 2020.

\bibitem[Kim et~al.(2021)Kim, Son, and Kim]{kim2021vilt}
Wonjae Kim, Bokyung Son, and Ildoo Kim.
\newblock Vilt: Vision-and-language transformer without convolution or region supervision.
\newblock In \emph{International conference on machine learning}, pages 5583--5594. PMLR, 2021.

\bibitem[Kovaleva et~al.(2019)Kovaleva, Romanov, Rogers, and Rumshisky]{kovaleva2019revealing}
Olga Kovaleva, Alexey Romanov, Anna Rogers, and Anna Rumshisky.
\newblock Revealing the dark secrets of bert.
\newblock \emph{arXiv preprint arXiv:1908.08593}, 2019.

\bibitem[Lau et~al.(2018)Lau, Gayen, Ben~Abacha, and Demner-Fushman]{lau2018dataset}
Jason~J Lau, Soumya Gayen, Asma Ben~Abacha, and Dina Demner-Fushman.
\newblock A dataset of clinically generated visual questions and answers about radiology images.
\newblock \emph{Scientific data}, 5\penalty0 (1):\penalty0 1--10, 2018.

\bibitem[Lee et~al.()Lee, Cho, and Kang]{leemixout}
Cheolhyoung Lee, Kyunghyun Cho, and Wanmo Kang.
\newblock Mixout: Effective regularization to finetune large-scale pretrained language models.
\newblock In \emph{International Conference on Learning Representations}.

\bibitem[Lee et~al.(2019)Lee, Tang, and Lin]{lee2019would}
Jaejun Lee, Raphael Tang, and Jimmy Lin.
\newblock What would elsa do? freezing layers during transformer fine-tuning.
\newblock \emph{arXiv preprint arXiv:1911.03090}, 2019.

\bibitem[Lee et~al.(2018)Lee, Ajanthan, and Torr]{lee2018snip}
Namhoon Lee, Thalaiyasingam Ajanthan, and Philip~HS Torr.
\newblock Snip: Single-shot network pruning based on connection sensitivity.
\newblock \emph{arXiv preprint arXiv:1810.02340}, 2018.

\bibitem[Lee et~al.(2023)Lee, Zhang, and Avestimehr]{lee2023layer}
Sunwoo Lee, Tuo Zhang, and A~Salman Avestimehr.
\newblock Layer-wise adaptive model aggregation for scalable federated learning.
\newblock In \emph{Proceedings of the AAAI Conference on Artificial Intelligence}, pages 8491--8499, 2023.

\bibitem[Lee et~al.(2022)Lee, Chen, Tajwar, Kumar, Yao, Liang, and Finn]{lee2022surgical}
Yoonho Lee, Annie~S Chen, Fahim Tajwar, Ananya Kumar, Huaxiu Yao, Percy Liang, and Chelsea Finn.
\newblock Surgical fine-tuning improves adaptation to distribution shifts.
\newblock \emph{arXiv preprint arXiv:2210.11466}, 2022.

\bibitem[Lester et~al.()Lester, Al-Rfou, and Constant]{lester2021power}
Brian Lester, Rami Al-Rfou, and Noah Constant.
\newblock The power of scale for parameter-efficient prompt tuning.
\newblock \emph{arXiv preprint arXiv:2104.08691}.

\bibitem[Li et~al.(2020{\natexlab{a}})Li, Kong, Zhang, Li, Li, Liu, and Ding]{li2020efficient}
Bingbing Li, Zhenglun Kong, Tianyun Zhang, Ji Li, Zhengang Li, Hang Liu, and Caiwen Ding.
\newblock Efficient transformer-based large scale language representations using hardware-friendly block structured pruning.
\newblock \emph{arXiv preprint arXiv:2009.08065}, 2020{\natexlab{a}}.

\bibitem[Li et~al.(2021)Li, Selvaraju, Gotmare, Joty, Xiong, and Hoi]{li2021align}
Junnan Li, Ramprasaath Selvaraju, Akhilesh Gotmare, Shafiq Joty, Caiming Xiong, and Steven Chu~Hong Hoi.
\newblock Align before fuse: Vision and language representation learning with momentum distillation.
\newblock \emph{Advances in neural information processing systems}, 34:\penalty0 9694--9705, 2021.

\bibitem[Li et~al.(2023)Li, Li, Savarese, and Hoi]{li2023blip}
Junnan Li, Dongxu Li, Silvio Savarese, and Steven Hoi.
\newblock Blip-2: Bootstrapping language-image pre-training with frozen image encoders and large language models.
\newblock In \emph{International conference on machine learning}, pages 19730--19742. PMLR, 2023.

\bibitem[Li et~al.(2020{\natexlab{b}})Li, Sahu, Talwalkar, and Smith]{li2020federated}
Tian Li, Anit~Kumar Sahu, Ameet Talwalkar, and Virginia Smith.
\newblock Federated learning: Challenges, methods, and future directions.
\newblock \emph{IEEE signal processing magazine}, 37\penalty0 (3):\penalty0 50--60, 2020{\natexlab{b}}.

\bibitem[Li and Liang(2021)]{li2021prefix}
Xiang~Lisa Li and Percy Liang.
\newblock Prefix-tuning: Optimizing continuous prompts for generation.
\newblock \emph{arXiv preprint arXiv:2101.00190}, 2021.

\bibitem[Lian et~al.(2022)Lian, Zhou, Feng, and Wang]{lian2022scaling}
Dongze Lian, Daquan Zhou, Jiashi Feng, and Xinchao Wang.
\newblock Scaling \& shifting your features: A new baseline for efficient model tuning.
\newblock \emph{arXiv preprint arXiv:2210.08823}, 2022.

\bibitem[Liu et~al.(2021)Liu, Zhan, Xu, Ma, Yang, and Wu]{liu2021slake}
Bo Liu, Li-Ming Zhan, Li Xu, Lin Ma, Yan Yang, and Xiao-Ming Wu.
\newblock Slake: A semantically-labeled knowledge-enhanced dataset for medical visual question answering.
\newblock In \emph{2021 IEEE 18th International Symposium on Biomedical Imaging (ISBI)}, pages 1650--1654. IEEE, 2021.

\bibitem[Liu et~al.(2024{\natexlab{a}})Liu, Li, Li, and Lee]{liu2024improved}
Haotian Liu, Chunyuan Li, Yuheng Li, and Yong~Jae Lee.
\newblock Improved baselines with visual instruction tuning.
\newblock In \emph{Proceedings of the IEEE/CVF Conference on Computer Vision and Pattern Recognition}, pages 26296--26306, 2024{\natexlab{a}}.

\bibitem[Liu et~al.(2024{\natexlab{b}})Liu, Shen, and Yang]{liu2024federated}
Renpu Liu, Cong Shen, and Jing Yang.
\newblock Federated representation learning in the under-parameterized regime.
\newblock \emph{arXiv preprint arXiv:2406.04596}, 2024{\natexlab{b}}.

\bibitem[Lu et~al.(2022)Lu, Wang, Chen, Qin, Xu, Dimitriadis, and Qin]{lu2022personalized}
Wang Lu, Jindong Wang, Yiqiang Chen, Xin Qin, Renjun Xu, Dimitrios Dimitriadis, and Tao Qin.
\newblock Personalized federated learning with adaptive batchnorm for healthcare.
\newblock \emph{IEEE Transactions on Big Data}, 2022.

\bibitem[Lubana and Dick()]{lubanagradient}
Ekdeep~Singh Lubana and Robert~P Dick.
\newblock A gradient flow framework for analyzing network pruning.
\newblock In \emph{International Conference on Learning Representations}.

\bibitem[McMahan et~al.(2017)McMahan, Moore, Ramage, Hampson, and y~Arcas]{mcmahan2017communication}
Brendan McMahan, Eider Moore, Daniel Ramage, Seth Hampson, and Blaise~Aguera y Arcas.
\newblock Communication-efficient learning of deep networks from decentralized data.
\newblock In \emph{Artificial intelligence and statistics}. PMLR, 2017.

\bibitem[Nguyen et~al.(2022)Nguyen, Malik, Sanjabi, and Rabbat]{nguyen2022begin}
John Nguyen, Kshitiz Malik, Maziar Sanjabi, and Michael Rabbat.
\newblock Where to begin? exploring the impact of pre-training and initialization in federated learning.
\newblock \emph{arXiv preprint arXiv:2206.15387}, 4, 2022.

\bibitem[Pillutla et~al.(2022)Pillutla, Malik, Mohamed, Rabbat, Sanjabi, and Xiao]{pillutla2022federated}
Krishna Pillutla, Kshitiz Malik, Abdel-Rahman Mohamed, Mike Rabbat, Maziar Sanjabi, and Lin Xiao.
\newblock Federated learning with partial model personalization.
\newblock In \emph{International Conference on Machine Learning}, pages 17716--17758. PMLR, 2022.

\bibitem[Rachwan et~al.(2022)Rachwan, Z{\"u}gner, Charpentier, Geisler, Ayle, and G{\"u}nnemann]{rachwan2022winning}
John Rachwan, Daniel Z{\"u}gner, Bertrand Charpentier, Simon Geisler, Morgane Ayle, and Stephan G{\"u}nnemann.
\newblock Winning the lottery ahead of time: Efficient early network pruning.
\newblock In \emph{International Conference on Machine Learning}, pages 18293--18309. PMLR, 2022.

\bibitem[Rahaman et~al.(2019)Rahaman, Baratin, Arpit, Draxler, Lin, Hamprecht, Bengio, and Courville]{rahaman2019spectral}
Nasim Rahaman, Aristide Baratin, Devansh Arpit, Felix Draxler, Min Lin, Fred Hamprecht, Yoshua Bengio, and Aaron Courville.
\newblock On the spectral bias of neural networks.
\newblock In \emph{International conference on machine learning}, pages 5301--5310. PMLR, 2019.

\bibitem[Rebuffi et~al.(2018)Rebuffi, Bilen, and Vedaldi]{rebuffi2018efficient}
Sylvestre-Alvise Rebuffi, Hakan Bilen, and Andrea Vedaldi.
\newblock Efficient parametrization of multi-domain deep neural networks.
\newblock In \emph{Proceedings of the IEEE Conference on Computer Vision and Pattern Recognition}, 2018.

\bibitem[R{\"u}ckl{\'e} et~al.(2020)R{\"u}ckl{\'e}, Geigle, Glockner, Beck, Pfeiffer, Reimers, and Gurevych]{ruckle2020adapterdrop}
Andreas R{\"u}ckl{\'e}, Gregor Geigle, Max Glockner, Tilman Beck, Jonas Pfeiffer, Nils Reimers, and Iryna Gurevych.
\newblock Adapterdrop: On the efficiency of adapters in transformers.
\newblock \emph{arXiv preprint arXiv:2010.11918}, 2020.

\bibitem[Saha et~al.(2023)Saha, Mishra, and Noble]{saha2023rethinking}
Pramit Saha, Divyanshu Mishra, and J~Alison Noble.
\newblock Rethinking semi-supervised federated learning: How to co-train fully-labeled and fully-unlabeled client imaging data.
\newblock In \emph{International Conference on Medical Image Computing and Computer-Assisted Intervention}, pages 414--424. Springer, 2023.

\bibitem[Saha et~al.(2024{\natexlab{a}})Saha, Mishra, Wagner, Kamnitsas, and Noble]{saha2024examiningmodalityincongruitymultimodal}
Pramit Saha, Divyanshu Mishra, Felix Wagner, Konstantinos Kamnitsas, and J.~Alison Noble.
\newblock Examining modality incongruity in multimodal federated learning for medical vision and language-based disease detection, 2024{\natexlab{a}}.

\bibitem[Saha et~al.(2024{\natexlab{b}})Saha, Mishra, Wagner, Kamnitsas, and Noble]{saha2024fedpia}
Pramit Saha, Divyanshu Mishra, Felix Wagner, Konstantinos Kamnitsas, and J~Alison Noble.
\newblock Fedpia--permuting and integrating adapters leveraging wasserstein barycenters for finetuning foundation models in multi-modal federated learning.
\newblock \emph{arXiv preprint arXiv:2412.14424}, 2024{\natexlab{b}}.

\bibitem[Shen et~al.(2021)Shen, Liu, Qin, Savvides, and Cheng]{shen2021partial}
Zhiqiang Shen, Zechun Liu, Jie Qin, Marios Savvides, and Kwang-Ting Cheng.
\newblock Partial is better than all: Revisiting fine-tuning strategy for few-shot learning.
\newblock In \emph{Proceedings of the AAAI conference on artificial intelligence}, pages 9594--9602, 2021.

\bibitem[Shi et~al.(2024)Shi, Chen, Dong, Yang, Li, Wang, Dick, Lv, Zhao, Yang, et~al.]{shi2024train}
Yubin Shi, Yixuan Chen, Mingzhi Dong, Xiaochen Yang, Dongsheng Li, Yujiang Wang, Robert Dick, Qin Lv, Yingying Zhao, Fan Yang, et~al.
\newblock Train faster, perform better: modular adaptive training in over-parameterized models.
\newblock \emph{Advances in Neural Information Processing Systems}, 36, 2024.

\bibitem[Sun et~al.(2022)Sun, Mendieta, Yang, and Chen]{sun2022exploring}
Guangyu Sun, Matias Mendieta, Taojiannan Yang, and Chen Chen.
\newblock Exploring parameter-efficient fine-tuning for improving communication efficiency in federated learning.
\newblock 2022.

\bibitem[Sung et~al.(2021)Sung, Nair, and Raffel]{sung2021training}
Yi-Lin Sung, Varun Nair, and Colin~A Raffel.
\newblock Training neural networks with fixed sparse masks.
\newblock \emph{Advances in Neural Information Processing Systems}, 34:\penalty0 24193--24205, 2021.

\bibitem[Tamirisa et~al.(2024)Tamirisa, Xie, Bao, Zhou, Arel, and Shamsian]{tamirisa2024fedselect}
Rishub Tamirisa, Chulin Xie, Wenxuan Bao, Andy Zhou, Ron Arel, and Aviv Shamsian.
\newblock Fedselect: Personalized federated learning with customized selection of parameters for fine-tuning.
\newblock In \emph{Proceedings of the IEEE/CVF Conference on Computer Vision and Pattern Recognition}, pages 23985--23994, 2024.

\bibitem[Tan et~al.(2022)Tan, Long, Liu, Zhou, Lu, Jiang, and Zhang]{tan2022fedproto}
Yue Tan, Guodong Long, Lu Liu, Tianyi Zhou, Qinghua Lu, Jing Jiang, and Chengqi Zhang.
\newblock Fedproto: Federated prototype learning across heterogeneous clients.
\newblock In \emph{Proceedings of the AAAI Conference on Artificial Intelligence}, pages 8432--8440, 2022.

\bibitem[Tanaka et~al.(2020)Tanaka, Kunin, Yamins, and Ganguli]{tanaka2020pruning}
Hidenori Tanaka, Daniel Kunin, Daniel~L Yamins, and Surya Ganguli.
\newblock Pruning neural networks without any data by iteratively conserving synaptic flow.
\newblock \emph{Advances in neural information processing systems}, 33:\penalty0 6377--6389, 2020.

\bibitem[Touvron et~al.(2022)Touvron, Cord, El-Nouby, Verbeek, and J{\'e}gou]{touvron2022three}
Hugo Touvron, Matthieu Cord, Alaaeldin El-Nouby, Jakob Verbeek, and Herv{\'e} J{\'e}gou.
\newblock Three things everyone should know about vision transformers.
\newblock In \emph{Computer Vision--ECCV 2022: 17th European Conference, Tel Aviv, Israel, October 23--27, 2022, Proceedings, Part XXIV}. Springer, 2022.

\bibitem[Van~Laarhoven et~al.(1987)Van~Laarhoven, Aarts, van Laarhoven, and Aarts]{van1987simulated}
Peter~JM Van~Laarhoven, Emile~HL Aarts, Peter~JM van Laarhoven, and Emile~HL Aarts.
\newblock \emph{Simulated annealing}.
\newblock Springer, 1987.

\bibitem[Wagner et~al.(2023)Wagner, Li, Saha, and Kamnitsas]{wagner2023post}
Felix Wagner, Zeju Li, Pramit Saha, and Konstantinos Kamnitsas.
\newblock Post-deployment adaptation with access to source data via federated learning and source-target remote gradient alignment.
\newblock In \emph{International Workshop on Machine Learning in Medical Imaging}, pages 253--263. Springer, 2023.

\bibitem[Wagner et~al.(2024)Wagner, Xu, Saha, Liang, Whitehouse, Menon, Newcombe, Voets, Noble, and Kamnitsas]{wagner2024feasibility}
Felix Wagner, Wentian Xu, Pramit Saha, Ziyun Liang, Daniel Whitehouse, David Menon, Virginia Newcombe, Natalie Voets, J~Alison Noble, and Konstantinos Kamnitsas.
\newblock Feasibility of federated learning from client databases with different brain diseases and mri modalities.
\newblock \emph{arXiv preprint arXiv:2406.11636}, 2024.

\bibitem[Wang et~al.(2020{\natexlab{a}})Wang, Zhang, and Grosse]{wang2020picking}
Chaoqi Wang, Guodong Zhang, and Roger Grosse.
\newblock Picking winning tickets before training by preserving gradient flow.
\newblock \emph{arXiv preprint arXiv:2002.07376}, 2020{\natexlab{a}}.

\bibitem[Wang et~al.(2020{\natexlab{b}})Wang, Liu, Liang, Joshi, and Poor]{wang2020tackling}
Jianyu Wang, Qinghua Liu, Hao Liang, Gauri Joshi, and H~Vincent Poor.
\newblock Tackling the objective inconsistency problem in heterogeneous federated optimization.
\newblock \emph{Advances in neural information processing systems}, 33:\penalty0 7611--7623, 2020{\natexlab{b}}.

\bibitem[Wen et~al.(2022)Wen, Jeon, and Huang]{wen2022federated}
Dingzhu Wen, Ki-Jun Jeon, and Kaibin Huang.
\newblock Federated dropout—a simple approach for enabling federated learning on resource constrained devices.
\newblock \emph{IEEE wireless communications letters}, 11\penalty0 (5):\penalty0 923--927, 2022.

\bibitem[Xu et~al.(2023)Xu, Tong, and Huang]{xu2023personalized}
Jian Xu, Xinyi Tong, and Shao-Lun Huang.
\newblock Personalized federated learning with feature alignment and classifier collaboration.
\newblock \emph{arXiv preprint arXiv:2306.11867}, 2023.

\bibitem[Xu et~al.(2021)Xu, Luo, Zhang, Tan, Chang, Huang, and Huang]{xu2021raise}
Runxin Xu, Fuli Luo, Zhiyuan Zhang, Chuanqi Tan, Baobao Chang, Songfang Huang, and Fei Huang.
\newblock Raise a child in large language model: Towards effective and generalizable fine-tuning.
\newblock \emph{arXiv preprint arXiv:2109.05687}, 2021.

\bibitem[Yang et~al.(2024{\natexlab{a}})Yang, Su, Li, and Xue]{yang2024exploring}
Mingzhao Yang, Shangchao Su, Bin Li, and Xiangyang Xue.
\newblock Exploring one-shot semi-supervised federated learning with pre-trained diffusion models.
\newblock In \emph{Proceedings of the AAAI Conference on Artificial Intelligence}, 2024{\natexlab{a}}.

\bibitem[Yang et~al.(2024{\natexlab{b}})Yang, Huang, and Ye]{yang2024fedas}
Xiyuan Yang, Wenke Huang, and Mang Ye.
\newblock Fedas: Bridging inconsistency in personalized federated learning.
\newblock In \emph{Proceedings of the IEEE/CVF Conference on Computer Vision and Pattern Recognition}, pages 11986--11995, 2024{\natexlab{b}}.

\bibitem[Yu et~al.(2023)Yu, Mu{\~n}oz, and Jannesari]{yu2023federated}
Sixing Yu, J~Pablo Mu{\~n}oz, and Ali Jannesari.
\newblock Federated foundation models: Privacy-preserving and collaborative learning for large models.
\newblock \emph{arXiv preprint arXiv:2305.11414}, 2023.

\bibitem[Zaken et~al.(2021)Zaken, Ravfogel, and Goldberg]{zaken2021bitfit}
Elad~Ben Zaken, Shauli Ravfogel, and Yoav Goldberg.
\newblock Bitfit: Simple parameter-efficient fine-tuning for transformer-based masked language-models.
\newblock \emph{arXiv preprint arXiv:2106.10199}, 2021.

\bibitem[Zhang et~al.(2023{\natexlab{a}})Zhang, Hua, Wang, Song, Xue, Ma, and Guan]{zhang2023fedala}
Jianqing Zhang, Yang Hua, Hao Wang, Tao Song, Zhengui Xue, Ruhui Ma, and Haibing Guan.
\newblock Fedala: Adaptive local aggregation for personalized federated learning.
\newblock In \emph{Proceedings of the AAAI Conference on Artificial Intelligence}, pages 11237--11244, 2023{\natexlab{a}}.

\bibitem[Zhang et~al.(2024)Zhang, Vahidian, Kuo, Li, Zhang, Yu, Wang, and Chen]{zhang2024towards}
Jianyi Zhang, Saeed Vahidian, Martin Kuo, Chunyuan Li, Ruiyi Zhang, Tong Yu, Guoyin Wang, and Yiran Chen.
\newblock Towards building the federatedgpt: Federated instruction tuning.
\newblock In \emph{ICASSP 2024-2024 IEEE International Conference on Acoustics, Speech and Signal Processing (ICASSP)}. IEEE, 2024.

\bibitem[Zhang et~al.(2023{\natexlab{b}})Zhang, Ding, Qi, Zhu, Long, and Zhou]{zhang2023crash}
Kaiyan Zhang, Ning Ding, Biqing Qi, Xuekai Zhu, Xinwei Long, and Bowen Zhou.
\newblock Crash: Clustering, removing, and sharing enhance fine-tuning without full large language model.
\newblock \emph{arXiv preprint arXiv:2310.15477}, 2023{\natexlab{b}}.

\bibitem[Zhang et~al.(2022)Zhang, Shen, Ding, Tao, and Duan]{zhang2022fine}
Lin Zhang, Li Shen, Liang Ding, Dacheng Tao, and Ling-Yu Duan.
\newblock Fine-tuning global model via data-free knowledge distillation for non-iid federated learning.
\newblock In \emph{Proceedings of the IEEE/CVF conference on computer vision and pattern recognition}, pages 10174--10183, 2022.

\bibitem[Zhang et~al.(2020)Zhang, Sapra, Fidler, Yeung, and Alvarez]{zhang2020personalized}
Michael Zhang, Karan Sapra, Sanja Fidler, Serena Yeung, and Jose~M Alvarez.
\newblock Personalized federated learning with first order model optimization.
\newblock \emph{arXiv preprint arXiv:2012.08565}, 2020.

\bibitem[Zhao et~al.(2024)Zhao, Sun, Shen, Yu, Kong, Wang, and Lin]{zhao2024pruning}
Pu Zhao, Fei Sun, Xuan Shen, Pinrui Yu, Zhenglun Kong, Yanzhi Wang, and Xue Lin.
\newblock Pruning foundation models for high accuracy without retraining.
\newblock \emph{arXiv preprint arXiv:2410.15567}, 2024.

\bibitem[Zhuang et~al.(2023)Zhuang, Chen, and Lyu]{zhuang2023foundation}
Weiming Zhuang, Chen Chen, and Lingjuan Lyu.
\newblock When foundation model meets federated learning: Motivations, challenges, and future directions.
\newblock \emph{arXiv preprint arXiv:2306.15546}, 2023.

\end{thebibliography}
}


\end{document}